\documentclass[letterpaper]{article} % DO NOT CHANGE THIS
\usepackage{aaai2026}  % DO NOT CHANGE THIS
\usepackage{times}  % DO NOT CHANGE THIS
\usepackage{helvet}  % DO NOT CHANGE THIS
\usepackage{courier}  % DO NOT CHANGE THIS
\usepackage[hyphens]{url}  % DO NOT CHANGE THIS
\usepackage{graphicx} % DO NOT CHANGE THIS
\usepackage{natbib}  % DO NOT CHANGE THIS AND DO NOT ADD ANY OPTIONS TO IT
\usepackage{caption} % DO NOT CHANGE THIS AND DO NOT ADD ANY OPTIONS TO IT
\usepackage{algorithm}
\usepackage{algorithmic}

\newtheorem{theorem}{Theorem}
\newtheorem{assumption}{Assumption}
\newtheorem{lemma}{Lemma}%[theorem]

\newtheorem{remark}{Remark}

\usepackage{newfloat}
\usepackage{listings}
\DeclareCaptionStyle{ruled}{labelfont=normalfont,labelsep=colon,strut=off} % DO NOT CHANGE THIS
\floatstyle{ruled}
\newfloat{listing}{tb}{lst}{}
\floatname{listing}{Listing}
\usepackage{amsmath}
\usepackage{amssymb}
\newenvironment{pf}[1][Proof]{%
  \par\vspace{0.5em}%
  \noindent\textbf{#1.}\ }{% 这里最后那个空格很重要
  \hfill$\square$\par\vspace{0.5em}%
}
\title{Scaling Law Analysis in Federated Learning: \\ 
How to Select the Optimal Model Size?}
\author {
    Xuanyu Chen\textsuperscript{\rm 1},
    Nan Yang\textsuperscript{\rm 1}$^{*}$,
    Shuai Wang\textsuperscript{\rm 2},
    Dong Yuan\textsuperscript{\rm 1}\thanks{Nan Yang and Dong Yuan are Corresponding authors. Contact email: n.yang@sydney.edu.au, dong.yuan@sydney.edu.au.},
}
\affiliations {
    \textsuperscript{\rm 1}The University of Sydney\\
    \textsuperscript{\rm 2}Northwestern Polytechnical University\\
}

\usepackage{bibentry}
\begin{document}

\maketitle

\begin{abstract}
The recent success of large language models (LLMs) has sparked a growing interest in training large-scale models. As the model size continues to scale, concerns are growing about the depletion of high-quality, well-curated training data. This has led practitioners to explore training approaches like Federated Learning (FL), which can leverage the abundant data on edge devices while maintaining privacy. However, the decentralization of training datasets in FL introduces challenges to scaling large models, a topic that remains under-explored. This paper fills this gap and provides qualitative insights on generalizing the previous model scaling experience to federated learning scenarios. Specifically, we derive a PAC-Bayes (Probably Approximately Correct Bayesian) upper bound for the generalization error of models trained with stochastic algorithms in federated settings and quantify the impact of distributed training data on the optimal model size by finding the analytic solution of model size that minimizes this bound. Our theoretical results demonstrate that the optimal model size has a negative power law relationship with the number of clients if the total training compute is unchanged. Besides, we also find that switching to FL with the same training compute will inevitably reduce the upper bound of generalization performance that the model can achieve through training, and that estimating the optimal model size in federated scenarios should depend on the average training compute across clients. Furthermore, we also empirically validate the correctness of our results with extensive training runs on different models, network settings, and datasets.
\end{abstract}

% Uncomment the following to link to your code, datasets, an extended version or similar.
% You must keep this block between (not within) the abstract and the main body of the paper.
\begin{links}
    % \link{Code}{https://aaai.org/example/code}
    % \link{Datasets}{https://aaai.org/example/datasets}
    \link{Extended Version}{https://aaai.org/example/extended-version}
    % \link{Appendix}{https://aaai.org/example/extended-version}
\end{links}

\section{Introduction}

In recent years, large-scale models with billions of parameters have been introduced and shown impressive performance on many tasks, such as Large Language Models (LLMs) \cite{brown2020language, rae2021scaling, thoppilan2022lamda}. Given the high cost of repeatedly training these massive models to achieve optimal results \cite{zhai2022scaling, tay2021scale}, researchers have developed scaling laws to estimate the compute-optimal model size, minimizing the need for repeated training, revealing that larger models can achieve better performance if trained with more data, as prescribed by these laws. For example, the Chinchilla model \cite{hoffmann2022training}, with 70 billion parameters, outperformed the Gopher model \cite{rae2021scaling}, which has 280 billion parameters, by following these laws to be trained on four times more data.

However, significantly expanding high-quality training data on centralized servers has long been a challenge \cite{muennighoff2024scaling}. In contrast, vast amounts of data are generated and stored in a distributed manner across society every day. A popular technique for utilizing this distributed data is Federated Learning (FL) \cite{mcmahan2017communication, abdulrahman2020survey}, where multiple clients collaboratively train a model while maintaining data privacy by preserving the training data locally. Notably, although it is less challenging to increase the size of training data in a federated learning scenario by scaling the network size (increasing the number of clients), the decentralization of training data will undoubtedly cause an impact on the scaling behavior of large-scale models summarized in previous studies. This motivates a question for practitioners seeking optimal training results: \textbf{\textit{How will the estimation of the optimal model size be affected when training large-scale models with distributed data using Federated Learning?}}

In this paper, we provide qualitative insights into generalizing existing model scaling laws to federated learning scenarios, supported by both theoretical analysis and empirical evidence. Theoretically, we model the Federated Learning process as a Stochastic Gradient Descent (SGD) optimization problem \cite{eon1998online, sutskever2013importance} over distributed data and establish a PAC-Bayes upper bound \cite{mcallester1998some, mcallester1999pac} for the generalization error of this learning algorithm. Using this generalization bound, we derive an analytic solution of the optimal model size minimizing this bound, quantifying the impact of distributed data on this optimal size. Empirically, we conduct several experiments to validate our theoretical findings. We pre-trained numerous models with varying parameter sizes using a popular architecture of auto-regressive transformer \cite{vaswani2017attention}, Vision Transformer \cite{dosovitskiy2020image, he2022masked} under centralized and federated settings. We then conducted linear probing on these models using two standard datasets, CIFAR-100 \cite{krizhevsky2009learning} and ImageNet \cite{deng2009imagenet}, and analyzed the final loss to demonstrate variations in optimal model size, which closely align with our theoretical predictions. 

In summary, the main contributions of this paper are shown below:

\begin{enumerate}
    \item We prove a PAC-Bayes upper bound for the generalization error of models trained by stochastic gradient descent optimization under federated settings and derive the analytic solution of the optimal model size under the convex settings of this bound. 
    \item By analyzing the analytic solution of the optimal model size, we obtain several qualitative insights on generalizing the scaling experience to federated scenarios: (i) the optimal model size has a negative power law relationship with the number of clients if the total training compute is unchanged; (ii) data decentralization in federated scenarios with a large number of clients will cause a drop in the upper bound of generalization performance that models can achieve through training; (iii) the optimal model size in federated scenarios can also be estimated from the average training compute allocated to each client.
    \item Furthermore, we conducted a sufficient number of training runs on models with different parameter sizes through centralized learning and federated learning, respectively. The empirical evidence validates the correctness of our theoretical results. 
\end{enumerate}

% The rest of this paper is organized as follows. We first review some related works. Then, we introduce some preliminaries. Next, we show our theoretical analyses, followed by our empirical validation results. Finally, we give a conclusion to the paper. The Appendix presents the details omitted from the main manuscript.

\section{Related Work}

\textbf{Scaling Law of LLMs.} Given the infeasibility of repeatedly training large language models (LLMs) with billions of parameters \cite{zhai2022scaling, tay2021scale}, researchers have developed various scaling laws to predict the relationship between the optimal model size and available training resources. Kaplan et al. were the first to observe a power-law relationship between model performance and model size \cite{kaplan2020scaling}, laying the foundation for subsequent works. Hoffman et al. revisited the problem under computational constraints and proposed the Chinchilla scaling law, which recommends equally scaling both model size and dataset size \cite{hoffmann2022training}. Recent studies have noted that the Chinchilla scaling law could deplete available training data, leading to the development of new scaling laws for data-constrained scenarios \cite{gadre2024language, muennighoff2024scaling}. However, these scaling laws are derived from empirical results in centralized training and may not directly apply to scenarios where training data is distributed. Our work addresses this gap by providing a theoretical analysis of the generalization bound and empirically validating the impact of data decentralization on scaling laws.

\noindent
\textbf{Federated Training of Large-scale Models.}
Federated Learning (FL) has garnered significant attention for enabling collaborative model training while preserving data privacy by keeping local data on clients \cite{mcmahan2017communication, abdulrahman2020survey, tang2022gossipfl}. Specifically, clients receive the global model from the central server, compute updates using their local data, and send these updates back to the server. The server aggregates the updates to refine the global model. This process is repeated until the model achieves the desired performance. With the rise of Large-scale models like LLMs, there has been a growing interest in applying FL to train these large-scale models \cite{chen2023federated}. Since clients generally have fewer computational resources than the server, most of these works follow the intuition to reduce the model size by tailoring the architecture of language models or freezing part of the model parameters during training \cite{xu2024fwdllm, wu2020fedmed}. However, few studies have explored the modified scaling behavior of language models in federated scenarios \cite{hilmkil2021scaling, ro2022scaling, shen2025will}, and those that have primarily offered observational insights based on empirical evidence. To address this theoretical gap, we model federated training as an SGD optimization problem over distributed data and quantify the impact on scaling by deriving an analytic solution for the optimal model size.

\noindent
\textbf{Generalization Bound for Stochastic Algorithms.} Stochastic gradient descent (SGD) \cite{eon1998online, sutskever2013importance} is a widely used optimization method in machine learning \cite{lecun1998gradient, hinton2006reducing, goodfellow2014generative, mcmahan2017communication, tang2022gossipfl}. Previous research has shown that the generalization performance of stochastic algorithms can be quantified using a PAC-Bayes upper bound \cite{he2019control, mou2018generalization, london2017pac, pensia2018generalization}, which is applied to explore various aspects, including algorithm convergence \cite{mou2018generalization, pensia2018generalization}, training stability \cite{zhu2024stability}, or strategy of tuning hyper-parameters \cite{he2019control}. The generalization bound also provides treatment for federated learning, helping several studies to propose new training frameworks to address the non-IID problem \cite{zhao2024federated} or model personalization \cite{boroujeni2024personalized, achituve2021personalized}, and other studies to figure out the impact of common parameters in federated scenarios on training results \cite{sefidgaran2024lessons}. In contrast, instead of proving similar generalization bounds for one of the two training regimes, our work focuses on comparing the generalization bounds of the stochastic algorithms in the federated settings with those in the centralized settings. The comparison results show us the impact of changing the training scenario on the optimal model size.

\section{Preliminaries}

\subsection{Bound for Generalization Error}
Formally, considering the hypothesis class of a model is $\Theta \subset \mathbb{R}^d$, machine learning algorithms aim to find the vector of model parameters $\theta \in \Theta $ that minimizes the expected risk $\mathcal{R(\theta)} = \mathbb{E}_{\xi \sim \mathcal{D}} F(\theta; \xi)$ where $d$ is the dimension of the parameter $\theta$, $F$ is the loss function, and $\mathcal{D}$ is the latent distribution of testing data. Suppose the output parameter $\theta$ follows a distribution $Q$, the expected risk in terms of $Q$ can be formulated as:
\begin{equation}
    \mathcal{R}(Q) = \mathbb{E}_{\theta \sim Q} \mathbb{E}_{\xi \sim \mathcal{D}} F(\theta; \xi).
\end{equation}
In practice, since $\mathcal{D}$ is not known in advance, the expected risk $\mathcal{R}$ is estimated by the empirical risk $\hat{\mathcal{R}}$ in terms of the latent distribution $\hat{\mathcal{D}}$ of the training data and is defined as:
\begin{equation}
    \hat{\mathcal{R}}(Q) = \mathbb{E}_{\theta \sim Q} \mathbb{E}_{\zeta \sim \hat{\mathcal{D}}} F(\theta; \zeta).
\end{equation}
The difference between $\mathcal{R}$ and $\hat{\mathcal{R}}$ is known as a generalization error, and the upper bound of the generalization error is usually used as a critical index to demonstrate the generalization ability of the training algorithm.

\subsection{SGD Optimization}
Stochastic Gradient Descent (SGD) is typically used to optimize the empirical risk $\hat{R}$. Consider a training dataset with size $m$, the mini-batch $\mathcal{S}$ of the training samples is equivalent to a subset of $S$ random indices that are independently and identically (i.i.d.) drawn from the index set $\{1, \ldots, m\}$. The SGD iteration can be formally defined as:
\begin{equation}
\label{eq_SGD_update}
\begin{aligned}
\theta(t+1) &=  \theta(t) - \eta \nabla_{\theta(t)} \hat{\mathcal{R}}(\theta(t)) \\
\quad &= \theta(t) - \eta \frac{1}{S} \sum\limits_{s\in \mathcal{S}} \nabla_{\theta(t)} F_s(\theta(t)).
\end{aligned}
\end{equation}
where $\nabla_{\theta_t} \hat{\mathcal{R}}(\theta_t)$ is the estimated gradient of empirical risk on mini-batch and $\eta$ is the learning rate.

\section{Theoretical Analysis}
In this section, we theoretically explore the impact of distributed data in federated learning on the optimal model size using a PAC-Bayes generalization bound for the stochastic algorithms. In particular, we establish the analytic solution of the optimal model size based on the proved bound, and compare the solutions between different scenarios to demonstrate several important insights. The detailed proof is omitted from this section and provided in Appendix A.

\subsection{Problem Setup}

To align with the standard settings of Federated Learning \cite{mcmahan2017communication}, we consider a distributed scenario consisting of $n$ clients and a central server that connects all clients. Each client $i \in \{1, \ldots, n\}$ possesses a local dataset $\mathcal{D}_i$, with the average dataset size denoted as $m = \frac{1}{n}\sum_{i=1}^n|\mathcal{D}_i|$. Thus, the total amount of data across all clients is $nm$. Suppose that training will be repeated for $T$ rounds, following the classical FL algorithm FedAvg \cite{mcmahan2017communication},  the training process at round $j\in \{1, \ldots, T\}$ can be expressed as:
% \begin{equation}
% \label{eq_client_train}
% \theta_{i}(j+1) = \Bar{\theta}(j) - \eta \nabla_{\Bar{\theta}(j)}\mathbb{E}_{\zeta_i \sim \mathcal{D}_i} F(\Bar{\theta}(j); \zeta_i),
% \end{equation}
\begin{equation}
\label{eq_fed_update}
\begin{aligned}
\Bar{\theta}(j+1) &= \frac{1}{n} \sum\limits_{i = 1}^n \theta_{i}(j+1) \\
&= \frac{1}{n} \sum\limits_{i = 1}^n (\Bar{\theta}(j) - \eta \nabla_{\Bar{\theta}(j)}\mathbb{E}_{\zeta_i \sim \mathcal{D}_i} F(\Bar{\theta}(j); \zeta_i)).
\end{aligned}
\end{equation}
% Eq.(\ref{eq_client_train}) shows the training of the global model $\Bar{\theta}(j)$ on client $i$ using its local dataset $\mathcal{D}_i$, and Eq.(\ref{eq_fed_update}) demonstrates the formal update of FL in each round by combining Eq.(\ref{eq_client_train}) with the model aggregation operation on the central server. 
Eq.(\ref{eq_fed_update}) demonstrates the formal update of FL in each round, with the first line denoting the model aggregation operation on the central server and the second line denoting the local training of the global model $\Bar{\theta}(j)$ on clients. Besides, since the training optimization is performed through SGD, we also define the batch size for local training as $S_{Fed} = k_{Fed}m \in \{1, \ldots, m\}$ where $\frac{1}{m} \leq k_{Fed} \leq 1$ and the number of local training epochs is $t$. Correspondingly, the baseline centralized scenario holds a dataset $\mathcal{D} = \bigcup_{i=1}^n \mathcal{D}_i $ of size $nm$, and the weights of the initial model in this scenario are the same as that in the federated scenario, i.e, $\{\theta(0) = \theta_i(0) | i \in n\}$.  The training of $\theta$ follows the SGD optimization described in Eq.(\ref{eq_SGD_update}), denoted as:
\begin{equation}
\label{eq_cen_update}
\theta(j+1) =  \theta(j) - \eta \nabla_{\theta(j)}\mathbb{E}_{\zeta \sim \mathcal{D}} F(\theta(j); \zeta),
\end{equation}
and is iterated for $\frac{T}{n}$ rounds to match the total training compute, which we define following the previous scaling law studies \cite{kaplan2020scaling,muennighoff2024scaling} as the total number of samples processed through training (i.e., dataset size times the number of training rounds). In each round, the model $\theta$ is trained using data from $\mathcal{D}$ for $t$ epochs with the batch size $S_{Cen} = k_{Cen}nm \in \{1, \ldots, nm\}$ where $\frac{1}{nm} \leq k_{Cen} \leq 1$. Furthermore, we have $k_{Fed}m \leq k_{Cen}nm$ due to more training data allocated to centralized settings in practice, leading to a generally larger batch size in use.

\subsection{A Generalization bound for Federated SGD}

To prove a PAC-Bayes generalization bound for the stochastic algorithms under federated settings, we first present some common assumptions aligned with the previous research \cite{stephan2017stochastic, he2019control}.

\begin{assumption}
\label{assum_C}
Considering that the stochastic gradient \( \hat{g}_s(\theta) = \nabla_{\theta(t)}\hat{\mathcal{R}}(\theta(t)) \) is computed as the sum of \( S \) independent gradients uniformly sampled from the training dataset, we assume that the gradient noise is Gaussian with covariance $\frac{1}{S} C(\theta)$, so $\hat{g}_s(\theta)$ can be approximated as
\begin{equation}
\hat{g}_s(\theta) \approx g(\theta) + \frac{1}{\sqrt{S}} \Delta g(\theta), \quad \Delta g(\theta) \sim \mathcal{N}(0, C(\theta)),
\end{equation}
where \( g(\theta) \) denotes the full gradient of the expected loss. We further assume that \( C(\theta) \) remains approximately constant with respect to $\theta$ and can be factorized into:
\begin{equation}
C(\theta) \approx C = BB^\top,
\end{equation}
where \( C \in \mathbb{R}^{d \times d} \) is symmetric and (semi) positive-definite. 
\end{assumption}

% We justify Assumption \ref{assum_C} by the central limit theorem when the training data size is substantially larger than the batch size. Since deep neural networks are typically trained on large-scale datasets in realistic cases, the Gaussian assumption about gradient noise is generally valid \cite{weinan2017proposal,stephan2017stochastic}. Also, the constant matrix $C$ can be justified when SGD iterates are confined to a small enough region around a local optimum of the loss, where the noise covariance does not vary significantly in that region.

We justify Assumption \ref{assum_C} by the central limit theorem: when the training data size is much larger than the batch size, the aggregated mini-batch gradients make the SGD update noise approximately Gaussian with near-stationary covariance. Since deep neural networks are typically trained on large-scale datasets, this Gaussian assumption generally holds and is standard in OU-based SGD analyses \cite{weinan2017proposal,stephan2017stochastic}. The constant matrix $C$ is further justified when SGD iterates are confined to a small enough region around a local optimum of the loss, where the noise covariance varies little.

\begin{assumption}
\label{assum_A}
Assuming the loss function $F(\theta)$ is smooth, when the stationary distribution of the iterates is confined to a local region near a minimum $\theta^*$, the loss gradient satisfies:
\begin{equation}
\nabla F(\theta) \approx A (\theta - \theta^*),
\end{equation} 
where \( A \in \mathbb{R}^{d \times d} \) is a constant (semi) positive-definite matrix representing the local Jacobian of the gradient field.
\end{assumption}

% Assumption \ref{assum_A} is generally valid when SGD converges to a low-variance quasi-stationary distribution near a deep local minimum, where the gradient noise is small compared to the average gradient. According to the fact that the exit time of a stochastic process is typically exponential in the height of the barriers between minima \cite{stephan2017stochastic}, local optima are very stable even in the presence of noise. Thus, SGD follows a relatively directed path toward the optimum. This assumption is also supported by empirical evidence (see p.1, Figures 1(a) and 1(b) and p.6, Figures 4(a) and 4(b) in \cite{li2018visualizing}). Moreover, this assumption can be extended to general cases through translation operations, which would not modify the geometry of the objective function and its associated generalization ability.

Assumption \ref{assum_A} is generally valid when SGD converges to a low-variance quasi-stationary distribution near a deep local minimum, where the gradient noise is small compared to the mean gradient. According to the fact that the exit time of a stochastic process is typically exponential in the barrier height between minima \cite{stephan2017stochastic}, local optima remain stable even in the presence of noise, and SGD thus follows a relatively directed path toward the optimum. This behavior is also supported by empirical observations in a prior study \cite{li2018visualizing}. Moreover, the assumption can be extended to general cases through a local translation. For example, translating around a stationary point $\theta^*$ with $\delta=\theta-\theta^*$ yields the Taylor form $\nabla F(\theta^*+\delta)=A\delta+O(\|\delta\|^2)$, assuming smoothness within a stable basin instead of global convexity.

Besides the above assumptions, we also need the formal definition of the PAC-Bayes upper bound to bound the generalization error. Following previous research \cite{mcallester1998some, mcallester1999pac}, we have:
\begin{lemma}
\label{lemma_pac_bayes}
For any positive real $\delta \in (0, 1)$, with probability at least $1 - \delta$ over a sample of size $N$, we have the following inequality for the distribution of the output hypothesis $Q$ and the prior $P$:
\begin{equation}
\label{eq_pac_bound}
\begin{aligned}
R(Q) \leq \hat{R}(Q) + \sqrt{\frac{\mathcal{D}(Q||P) + \log(\frac{1}{\delta})+\log(N)+2}{2N - 1}},
\end{aligned}
\end{equation}
where  $\mathcal{D}(Q||P)$ is the KL divergence between the distributions $Q$ and $P$ and is defined as: $\mathcal{D}(Q||P) = \mathbb{E}_{\theta \sim Q} \log(\frac{Q(\theta)}{P(\theta)})$.
\end{lemma}

Based on the two assumptions and Lemma \ref{lemma_pac_bayes}, we can prove the generalization bound for federated SGD below.

\begin{theorem}
\label{theorem_dec_pac_bound}
For any positive real $\delta \in (0, 1)$, with probability at least $1 - \delta$ over a distributed training data set with a total size $nm$ across all clients, we have the following inequality for the distribution $Q_{Fed}$ of the output hypothesis function of federated SGD:
\begin{equation}
\label{eq_fed_pac_bound}
\begin{aligned}
& R(Q_{Fed}) - \hat{R}(Q_{Fed})  \leq \\
& \sqrt{\frac{H_1 + H_2 - d + 2\log(\frac{1}{\delta}) + 2\log (nm) + 4}{4nm - 2}},
\end{aligned}
\end{equation}
where 
\begin{equation}
H_1 = -\log(\det(\Sigma_{Fed})), H_2 = \frac{T\eta}{2k_{Fed}m} \text{tr}(\Bar{C}\Bar{A}^{-1}),
\end{equation}
$d$ is the dimension of the parameter (the model size) and $\text{tr}(\Bar{C}\Bar{A}^{-1})$ is the trace of the product matrix $\Bar{C}\Bar{A}^{-1}$.
\end{theorem}

Apparently, it is hard to quantify the above generalization bound since the covariance matrix $\Sigma_{Fed}$ for the stationary distribution is not available for the training data. In order to be able to estimate the optimal model size using the generalization bound, we introduce the following assumption and study a special case of the generalization bound as in other papers \cite{he2019control, jastrzkebski2017three}.

\begin{assumption}
\label{assum_symmetric}
We assume that $A$ and $\Sigma$ are symmetric matrices satisfying $A\Sigma = \Sigma A$.  
\end{assumption}

Assumption \ref{assum_symmetric} implies that the local geometry around the global minimum and the stationary distribution is homogeneous across all dimensions of the parameter space, so $A$ and $\Sigma$ can be assumed to share eigenvectors. This common isotropic simplification, also adopted in prior papers \cite{he2019control, jastrzkebski2017three}, makes trace and determinant terms analytically solvable without affecting the scaling order of the bound. When Assumption \ref{assum_symmetric} also holds, we reformulate the property found in the proof of Theorem \ref{theorem_dec_pac_bound} and derive a new generalization bound as follows.
\begin{theorem}
\label{theorem_new_dec_pac_bound}
Under all the Assumptions of Theorem \ref{theorem_dec_pac_bound} and with Assumption \ref{assum_symmetric}, we have the following generalization bound for the stationary distribution of federated SGD:  
\begin{equation}
\label{dec_pac_bound}
\begin{aligned}
& R(Q_{Fed}) - \hat{R}(Q_{Fed}) \leq\\
&\sqrt{\frac{H_{Fed} + H_{Fed}^{'} - d + 2\log(\frac{1}{\delta}) + 2\log (nm) + 4}{4nm - 2}},
\end{aligned}
\end{equation}
where $H_{Fed} = d\log(\frac{2k_{Fed}m}{T\eta}) - \log(\det(\Bar{C}\Bar{A}^{-1}))$ and $H_{Fed}^{'} = \frac{T\eta}{2k_{Fed}m} \text{tr}(\Bar{C}\Bar{A}^{-1})$.
\end{theorem}

\subsection{Relationship between Two Optimal Model Sizes}
In the same way, we can prove the generalization bound for centralized SGD with the same amount of training data. 

\begin{lemma}
\label{lemma_cen_pac_bound}
Under all the assumptions of Theorem \ref{theorem_new_dec_pac_bound}, we have the following generalization bound for the stationary distribution of centralized SGD trained on the same amount of training data:  
\begin{equation}
\label{cen_pac_bound}
\begin{aligned}
& R(Q_{Cen}) - \hat{R}(Q_{Cen}) \leq \\
& \sqrt{\frac{H_{Cen} + H_{Cen}^{'} - d + 2\log(\frac{1}{\delta}) + 2\log (nm) + 4}{4nm - 2}}.
\end{aligned}
\end{equation}
where $H_{Cen} = d\log(\frac{2k_{Cen}n^2m}{T\eta}) - \log(\det(CA^{-1}))$ and $H_{Cen}^{'} = \frac{T\eta}{2k_{Cen}n^2m} \text{tr}(CA^{-1})$.
\end{lemma}

Generalization bounds provide an upper limit on an algorithm’s generalization error, with smaller bounds indicating better generalization performance. A natural criterion for selecting the optimal model size $d^*$ is the value of $d$ that minimizes this bound. While convexity is not guaranteed in general, empirical studies on scaling laws \cite{kaplan2020scaling, hoffmann2022training, muennighoff2024scaling} suggest an approximately convex relationship between model size and generalization performance. Based on this, we assume a locally convex regime and derive a closed-form approximation of $d^*$ using first-order conditions.

\begin{lemma}
\label{lemma_size_solution}
When all the above assumptions hold, the optimal model size under the output hypothesis function of federated SGD has the following analytic solution:
\begin{equation}
\label{eq_optimal_size_fed}
d_{Fed}^* = \frac{H_1 + H_2 + 8n\log(\frac{1}{\delta}) + 8n\log(nm) - \frac{4}{m} + 8n}{8n - \frac{2}{m} - 4n\log(\frac{2k_{Fed}m}{T\eta})}.
\end{equation}
where $H_1 = -4n\log((\det(\Bar{C}\Bar{A}^{-1}))$ and $H_2 = (\frac{4nT\eta}{k_{Fed}m} - \frac{T\eta}{k_{Fed}m^2}) \text{tr}(\Bar{C}\Bar{A}^{-1}))$. 

On the other hand, the optimal model size for centralized SGD has the following analytic solution:
\begin{equation}
\label{eq_optimal_size_cen}
d_{Cen}^* = \frac{\hat{H}_1 + \hat{H}_2 + 8n\log(\frac{1}{\delta}) + 8n\log(nm) - \frac{4}{m} + 8n}{8n - \frac{2}{m} - 4n\log(\frac{2k_{Cen}n^2m}{T\eta})}.
\end{equation}
where $\hat{H}_1 = -4n\log((\det(CA^{-1}))$ and $\hat{H}_2 = (\frac{4T\eta}{k_{Cen}nm} - \frac{T\eta}{k_{Cen}n^2m^2}) \text{tr}(CA^{-1}))$. 
\end{lemma}

The comparison between $d_{Fed}^*$ and $d_{Cen}^*$ reveals their relationship. However, we cannot quantify this relationship without knowing the connection between the average local distribution and the global distribution (i.e., $\text{tr}(\bar{C}\bar{A}^{-1}))$ vs $\text{tr}(CA^{-1}))$). Hence, we introduce another assumption.

\begin{assumption}
\label{assump_fair_comp}
Under the fair comparison condition that the same training dataset is used for both training scenarios, we assume that the local data distributions \( \mathcal{D}_1, \dots, \mathcal{D}_n \) across \( n \) clients of size \( m \) form a heterogeneous partition of the global dataset \( \mathcal{D} \) of size \( D = nm \). Hence, we have the following approximate relationships:
\begin{equation}
\Bar{A} \approx A + \Delta_A, \quad \Bar{C} \approx \frac{1}{n^{\gamma}} (C + \Delta_C),
\end{equation}
where \( \gamma > 1 \), and \( \Delta_A \), \( \Delta_C \) are deviation terms introduced by data heterogeneity. These deviations are assumed to be bounded in norm:
\begin{equation}
\|\Delta_A\| \leq \epsilon_A, \quad \|\Delta_C\| \leq \epsilon_C,
\end{equation}
where \( \epsilon_A, \epsilon_C \) grow with the non-i.i.d degree across clients.
\end{assumption}

% Assumption \ref{assump_fair_comp} reflects the realistic cases where client datasets are drawn from heterogeneous (non-i.i.d.) distributions and could be justified by the central limit theorem when the average data size $m$ across clients and the size of the global dataset $D$ are both large enough. While the centralized quantities $A$ and $C$ characterize curvature and noise under the full dataset, their decentralized counterparts $\Bar{A}$ and $\Bar{C}$ may deviate due to non-i.i.d. sampling. The inclusion of bounded deviation terms $\Delta_A$ and $\Delta_C$, whose magnitudes reflect the degree of data heterogeneity across clients, and the scaling variable $\gamma$ captures this variability while retaining analytical traceability for us to quantify the impact of non-i.i.d. distributions. Under this assumption, the relationship between the two optimal model sizes is shown below.

Assumption \ref{assump_fair_comp} reflects realistic FL scenarios where client data are drawn from heterogeneous (non-i.i.d.) distributions. It can be justified by the central limit theorem when both the average local data size $m$ and the global dataset $D$ are large. While the centralized matrices $A$ and $C$ describe curvature and gradient noise under the full dataset, their decentralized averages $\bar A$ and $\bar C$ deviate under non-i.i.d. sampling. Following the bounded-variance formulation in FL theory \cite{karimireddy2020scaffold}, we include deviation terms $\Delta_A$ and $\Delta_C$ whose magnitudes quantify heterogeneity, and introduce the scaling exponent $\gamma>1$ to capture how such heterogeneity amplifies the gap between local and global covariances while keeping the analysis analytically tractable. Under this assumption, the relationship between the two optimal model sizes is shown below.

\begin{theorem}
\label{theorem_size_relation}
When all the above assumptions hold, by comparing the optimal model size between the federated and centralized scenarios, we find that:
\begin{equation}
\lim_{T \to \infty} d_{Fed}^* = \frac{\rho}{n^{\gamma - 1}}d_{Cen}^*,  
\end{equation}
where $\rho = \frac{S_{Cen} \left(\text{tr}(CA^{-1}) + \text{tr}(\Delta_1)\right)}{S_{Fed} \text{tr}(CA^{-1})} > 0$ and $\Delta_1 = (CA^{-1}\Delta_A + \Delta_C(I+A^{-1}\Delta_A))A^{-1}$.
\end{theorem}

\begin{remark}
Since we have $\gamma > 1$, Theorem \ref{theorem_size_relation} shows $d_{Fed}^* < d_{Cen}^*$ and suggests the \textbf{first theoretical insight:} 
\begin{itemize}
    \item When transferring the training of large-scale models from centralized to federated scenarios with the same training compute, the optimal model size should be decreased, and the reduction ratio has a negative power law relationship with the number of clients.
\end{itemize}
\end{remark}

\subsection{Theoretical Evidence for Generalization Gap}
The above theoretical proofs demonstrate the generalization bound and the optimal model size under this bound for each training. In this subsection, we show that these proofs can also serve as an important theoretical basis for an empirical finding observed in many previous works. Specifically, models trained with distributed data are generally found to be inferior to the models trained with centralized data in performance \cite{lian2017can, sun2022adaptive}. According to the respective generalization bound and optimal model size, we derive the following theorem.

\begin{theorem}
\label{theorem_gap}
When all the above assumptions hold, we find the following inequality between the optimal generalization error of federated SGD and centralized SGD using the same training compute: 
\begin{equation} 
\label{eq_theorem_gap_1}
\lim_{T\to\infty}(\mathcal{G}_{Fed}^* - \mathcal{G}_{Cen}^*) > 0 
\end{equation}
when the number of clients $n$ satisfies the property: $n > \sqrt[\gamma - 1]{\rho}$. Here, $\mathcal{G}^*$ is the optimal generalization error computed with the optimal model size $d^*$.
\end{theorem}

\begin{remark}
The condition in Theorem \ref{theorem_gap} basically holds in practice, considering that realistic federated scenarios generally scale to a sufficiently large number of clients \cite{kairouz2021advances} (e.g., phones with user data, edge sensors, etc.) Also, it is expected that the value of $\rho$ would not be very large. In the ideal case where client data is i.i.d. and both training uses identical batch size, we have $\rho = 1$. Since $n \geq 2$ holds for any federated scenarios, the inequality will be trivially satisfied. Based on this result, we summarize our \textbf{second theoretical insight:}
\begin{itemize}
    \item If a federated scenario with a large number of clients is not allocated more data than the centralized scenario, data decentralization will lead to a definite gap between the optimal generalization performance achieved through FL and that under centralized settings, which underscores the challenges of FL.
\end{itemize}
\end{remark}

\subsection{Estimating Optimal Model Size by the Average Training Compute Between Clients}

Previous analyses have studied the optimal model size for federated SGD training using all local data from the clients. Notably, the total training compute for the federated SGD training is equal to the sum of the training compute allocated to each client. Therefore, we are also interested in the optimal model size at the local level and how it relates to the above results. By a similar proof, we derive the analytic solution of the optimal model size at the client level as follows. 

\begin{lemma}
\label{lemma_single_client_size}
When all the above assumptions hold, the optimal model size at the client level $d_{i}^*$ has the following analytic solution:
\begin{equation}
\label{eq_optimal_size_1c}
d_{i}^* = \frac{ H_1^{(i)} + H_2^{(i)} + 8\log(m) - \frac{4}{m} + 8}{8 - \frac{2}{m} - 4\log(\frac{2k_{i}m}{T\eta})},
\end{equation}
where $H_1^{(i)} = -4\log((\det(C_iA_i^{-1}))$ and $H_2^{(i)} = (\frac{4T\eta}{k_im} - \frac{T\eta}{k_im^2}) + 8\log(\frac{1}{\delta})-4\log((\det(C_iA_i^{-1}))$.
\end{lemma}
Then, considering that the local data on clients is heterogeneous, we use $\xi_{i}^C = C_i - \bar{C}$ and $\xi_{i}^A = A_i - \bar{A}$ to denote client variance in non-IID settings. Comparing $d_{i}^*$ with the optimal model size $d_{Fed}^*$ in FL across $n$ clients shows their relationship as follows.
\begin{theorem}
\label{theorem_client_size_relation}
When all the above assumptions hold, considering the same batch size $\{k_{Fed}m = k_{i}m | i \in n\}$, the following relation holds between the optimal model size $d_{i}^*$ on a single client and the optimal model size $d_{Fed}^*$ of FL across $n$ clients: 
\begin{equation}
\label{eq_single_client_relation}
\lim_{T\to\infty}d_{Fed}^* = \frac{\kappa}{n} \sum\limits_{i=1}^n d_i^*, 
\end{equation}
where $\kappa = \frac{\left(4m - \frac{1}{n}\right)\text{tr}(\bar{C}\bar{A}^{-1})}{\left(4m-1\right)\left(\text{tr}(\bar{C}\bar{A}^{-1}) + \text{tr}(\bar{\xi})\right)} > 0$ and $\bar{\xi} = \frac{1}{n}\sum_{i=1}^{n}((\bar{C}\bar{A}^{-1}\xi_i^A + \xi_i^C(I+\bar{A}^{-1}\xi_i^A))\bar{A}^{-1})$.
\end{theorem}
\begin{remark}
Since the existing scaling law suggests that the optimal model size relates to the data size \cite{kaplan2020scaling}, it is intuitive that the optimal model size in FL would be decided by the total data size across clients (i.e., $d_{Fed}^* \approx \sum_{i=1}^n d_i^*$). However, Theorem \ref{theorem_client_size_relation} demonstrates that this thought is incorrect. Eq.(\ref{eq_single_client_relation}) implies $d_{Fed}^* \approx \frac{1}{n} \sum_{i=1}^n d_i^*$, as $\frac{4m - \frac{1}{n}}{4m - 1} \approx 1$ and the average bias term $\text{tr}(\bar{\xi})$ is much smaller than $\text{tr}(\bar{C}\bar{A}^{-1})$. This highlights our \textbf{third theoretical insight:}
\begin{itemize}
    \item The optimal model size in FL is primarily determined by the average training compute per client, rather than the total compute across all clients or the number of clients.
\end{itemize}
\end{remark}

\section{Empirical Validation}

\subsection{Experiment Setup}
We conduct multiple experiments based on a popular model architecture, Vision Transformer (ViT) \cite{dosovitskiy2020image}. This architecture represents a dominant type of model in deep learning: Transformers \cite{vaswani2017attention} relying on the attention mechanism, which is frequently used for building large-scale models. Specifically, we build 10 different sizes of ViTs with parameters ranging from $11.62$ to $75.41$ million. These models are pre-trained on the Mini-Imagenet dataset \cite{vinyals2016matching}, which contains 60,000 images extracted from the ImageNet dataset \cite{deng2009imagenet}. We adopt the Masked Auto-encoder \cite{he2022masked} (MAE) approach to pre-train ViTs. To evaluate the effectiveness of pre-training, we conduct linear probing tests, which freeze the pre-trained weights in the backbone and only fine-tune the head layer \cite{he2016deep}. The linear probing accuracies of these models on two standard datasets (CIFAR-100 \cite{krizhevsky2009learning} and ImageNet \cite{deng2009imagenet}) are collected for analysis. We select the size of the model with the highest linear probing (LP) accuracy as the optimal model size.

For all experiments, we strictly follow the problem setup defined in the theoretical analysis. All training resources are kept the same between the centralized and federated scenarios, including the model, total training compute, and dataset. To simulate federated scenarios with $n$ clients and non-IID client data, we divide the training dataset into $n$ partitions by sampling the class priors of the Dirichlet distribution \cite{hsu2019measuring}. A more heterogeneous division can be made by specifying a smaller Dirichlet parameter $\alpha$ during sampling. We use $\alpha = 0.1$ by default.  Our codes for experiments were implemented using the PyTorch framework and executed on a server with 8 $\text{NVIDIA\textregistered}$ RTX A5000 GPUs. The details about the experiment and server settings are provided in Appendix B. 

\begin{figure}[t]
    \centering
    % 第一张图
    \includegraphics[width=\linewidth]{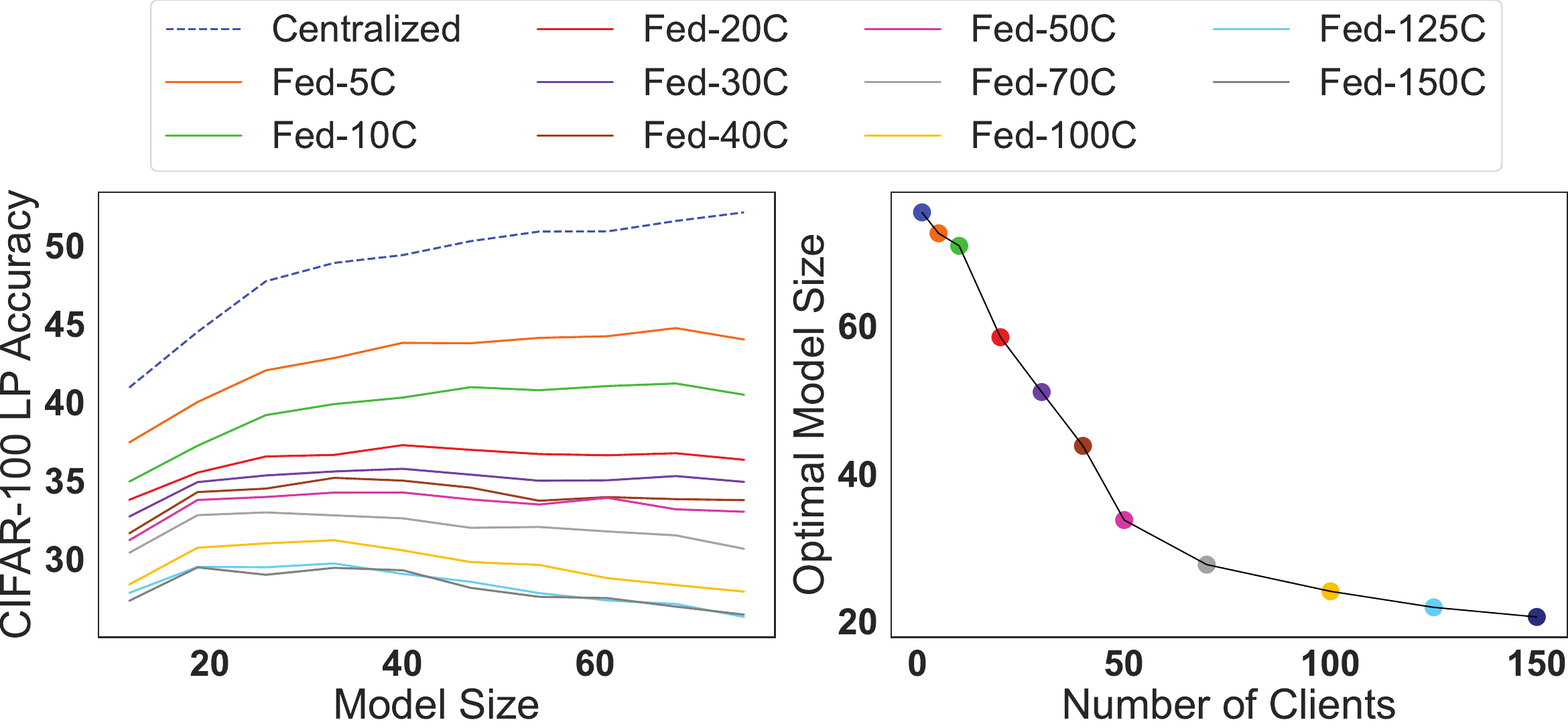}
    \caption*{\textbf{(a) CIFAR-100}}

    % 垂直间距
    \vspace{0.5em}

    % 第二张图
    \includegraphics[width=\linewidth]{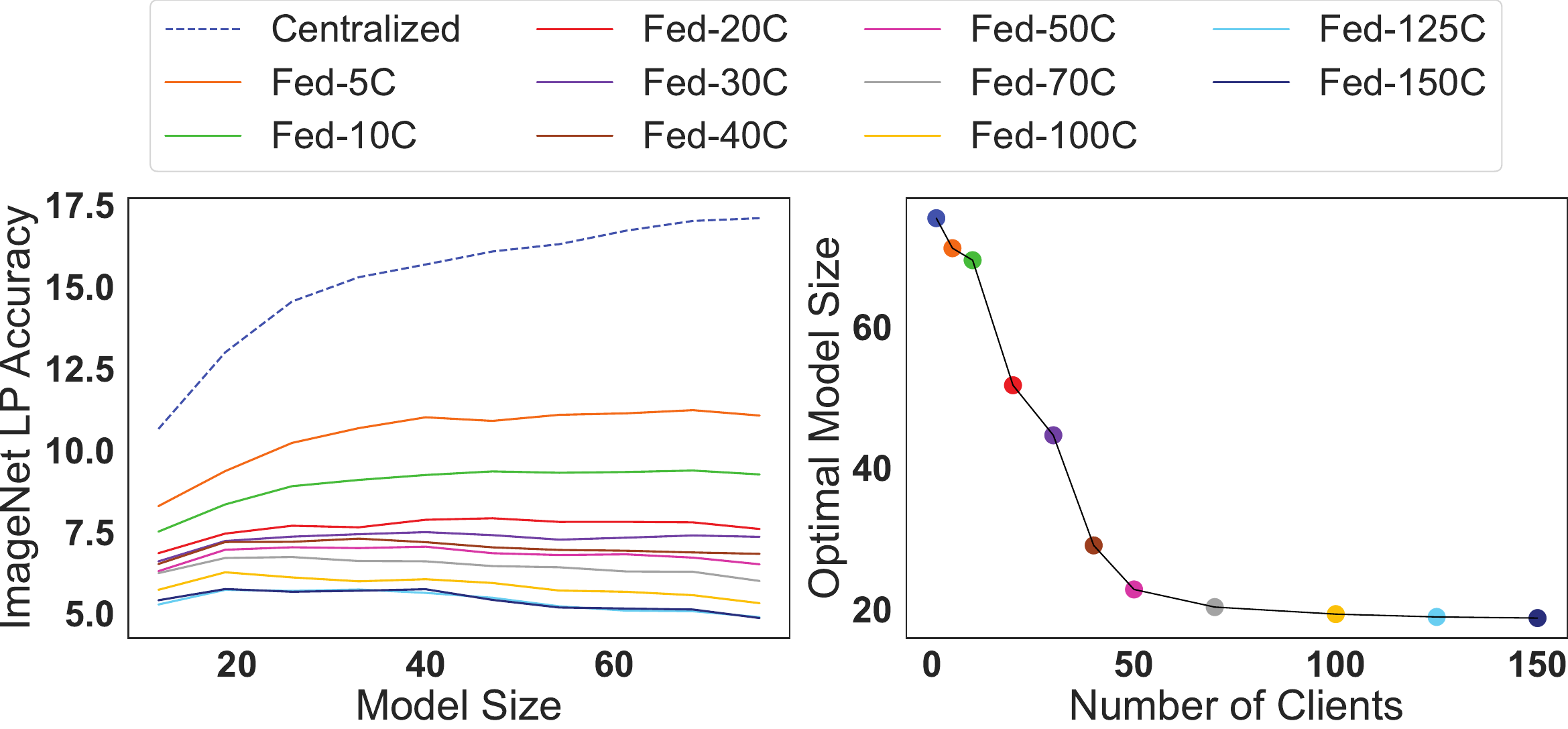}
    \caption*{\textbf{(b) ImageNet}}

    % 总标题
    \caption{\textbf{Impact of distributed data on the optimal model size of ViT.} 
    (Left) Curves of LP accuracy ($\%$) versus model size. Different lines represent FL scenarios with a different number of clients. 
    (Right) Curve of optimal model size versus the number of clients. Here, the centralized setting corresponds to the case $n=1$.}
\label{fig:combined_model_size}
\end{figure}

\subsection{Empirical Results}

% \textbf{Evidence for the first insight.} We investigate the impact of distributed data on the optimal model size by training models with the same training compute in both the centralized scenario and federated scenarios with a different number of clients. Figure \ref{fig:combined_model_size}(a-Left) shows the linear probing accuracies of ViTs with different sizes on the CIFAR-100 dataset in each scenario. Based on the highest accuracy, we find the optimal model size for each scenario and plot them in Figure \ref{fig:combined_model_size}(a-Right). The results clearly show that there is a negative power law relationship between the optimal model size and the number of clients, which validates our theoretical result in Theorem \ref{theorem_size_relation}. Besides, we have also collected the linear probing results of ViTs on the ImageNet dataset \cite{deng2009imagenet}. This dataset has approximately 1.2 million training images, which is much larger than the CIFAR-100 dataset. We use $10\%$ of the training data for linear probing and still observe similar empirical results, as shown in Figure \ref{fig:combined_model_size}(b). 

\textbf{Evidence for the First Insight.} We investigate the impact of distributed data on the optimal model size by training models with the same training compute in both the centralized scenario and federated scenarios with different numbers of clients. 
Figure~\ref{fig:combined_model_size}(a, Left) shows the linear probing accuracies of ViTs with different sizes on CIFAR-100 in each scenario. 
Based on the highest accuracy, we find the optimal model size for each scenario and plot them in Figure~\ref{fig:combined_model_size}(a, Right). 
The results clearly show a negative power-law relationship between the optimal model size and the number of clients, validating Theorem~\ref{theorem_size_relation}. 
Besides, we have also collected the linear probing results of ViTs on the ImageNet dataset \cite{deng2009imagenet}, which has around 1.2 million images. We use $10\%$ of the training samples for linear probing and still observe similar empirical results, as shown in Figure \ref{fig:combined_model_size}(b). To further validate the size relationship, we fit the log–log form of Theorem~\ref{theorem_size_relation} (i.e., $\log d^*_{Fed} = (1-\gamma)\log n + \log\rho + \log d^*_{Cen}$) by the measured optimal sizes in FL. This gives slopes in $[-0.28,-0.48]$ ($\gamma\in[1.3,1.5], \rho  \in[2.0,2.4]$) with coefficient of determination $R^2>0.88$, confirming a strong negative power law in $n$.

% \begin{table}[t!]
% \centering
% \footnotesize
% \caption{\textbf{Empirical fitting of Theorem \ref{theorem_size_relation} across models and datasets.} Slopes correspond to $(1-\gamma)$ estimated from log-log regression, and $R^2$ is the coefficient of determination. Higher $R^2$ (closer to 1) show a better fit.}
% \label{tab:scaling_fit}
% \begin{tabular}{l|cccc}
% \hline
% Setting & Slope & $\gamma$ & $\rho$ & $R^2$   \\
% \hline
% ViT + CIFAR-100 & $-0.420$  & $1.420$ & $2.366$ & $0.932$ \\
% ViT + ImageNet & $-0.475$  & $1.475$ & $2.406$ & $0.920$ \\
% ResNet + CIFAR-100 & $-0.282$ & $1.282$ & $0.902$ & $0.877$ \\
% \hline
% \end{tabular}
% \end{table}

\begin{figure}[t!]
    \centering
    % 第一张图
    \includegraphics[width=0.7\linewidth]{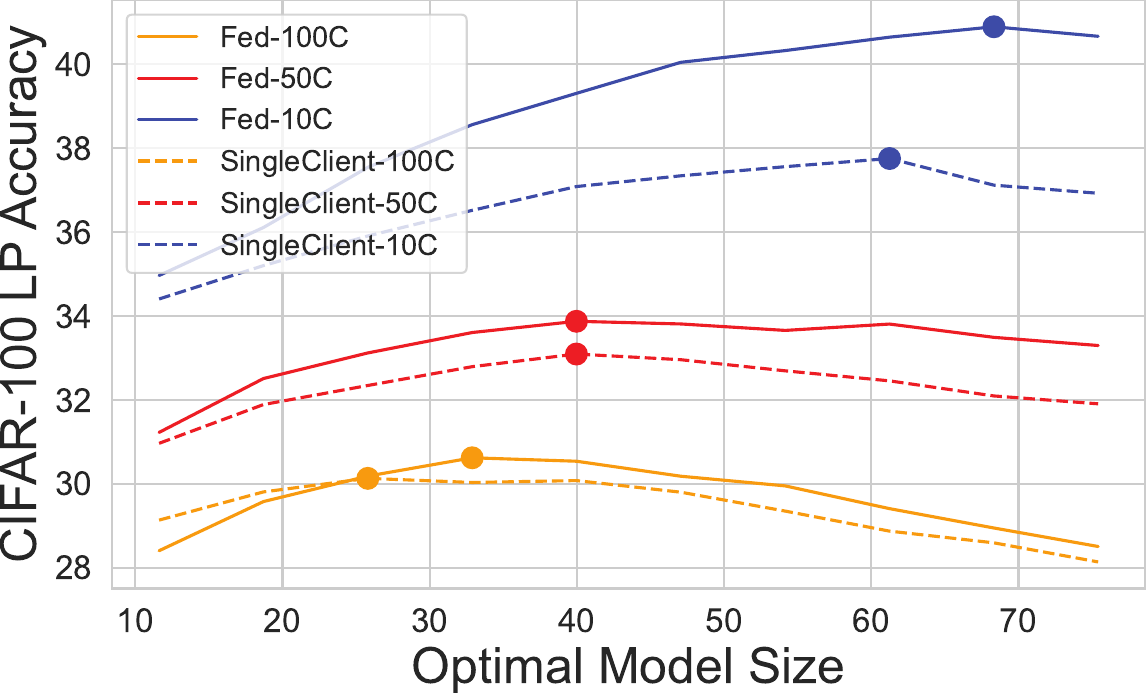}
    \caption*{\textbf{(a) CIFAR-100}}

    % 两图之间的竖直间距
    \vspace{0.5em}

    % 第二张图
    \includegraphics[width=0.7\linewidth]{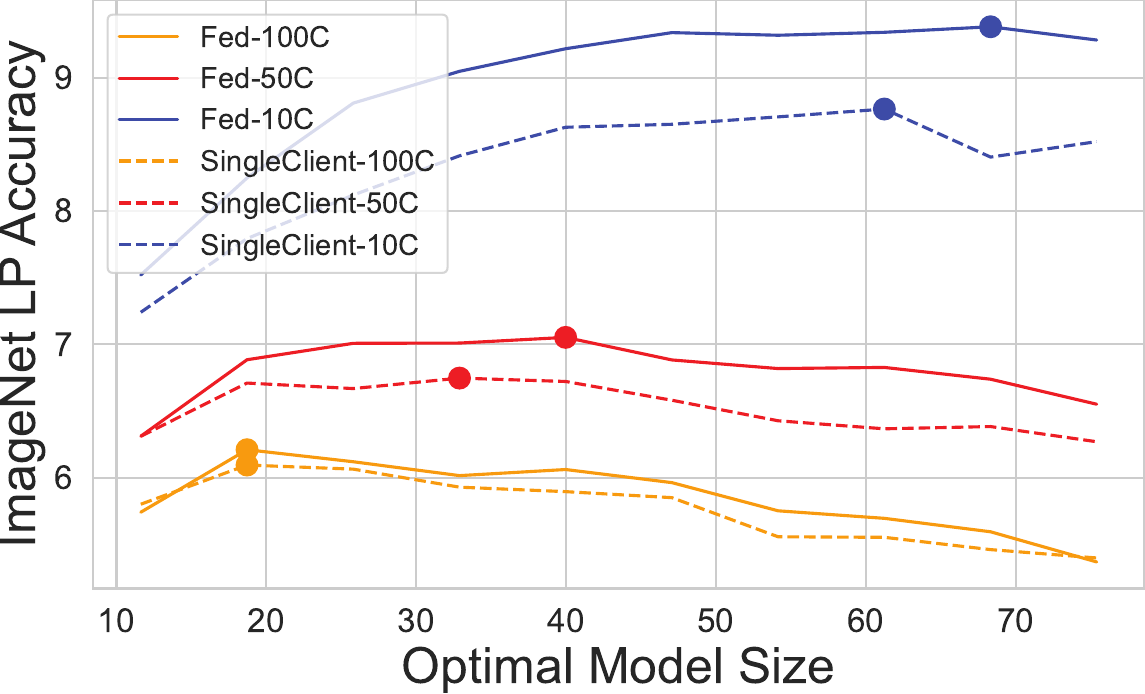}
    \caption*{\textbf{(b) ImageNet}}

    % 总标题
    \caption{\textbf{Comparison between the optimal model size across all clients and for a single client.} 
    The dots represent the highest accuracy of each line.}
    \label{fig:combined_single_client}
\end{figure}

\noindent
\textbf{Evidence for the Second Insight.} Figure \ref{fig:combined_model_size}(Left) also shows the impact of the number of clients on the linear probing accuracy of models. We note that when the federated scenario does not have an advantage on the size of training data (i.e., the total training data across clients equals the training data used in centralized training), the models trained by federated learning will be inferior to the models trained by centralized learning in terms of the generalization performance even if the number of clients is very small (a good example here is $n=5$). Moreover, the gap between the two training regimes broadens as the number of clients increases. These findings are consistent with our theoretical results in Theorem \ref{theorem_gap}.

\noindent
\textbf{Evidence for the Third Insight.} FL indirectly uses all training data from clients to train a global model by an iterative process of having multiple models trained on different clients using their local data and aggregating the parameters of these models on the server. In Figure \ref{fig:combined_single_client}, we train models using only local data from a single client and compute the average of $n$ sets of linear-probing accuracies from $n$ clients. These results are then compared with those of FL. We observe that the optimal model size is actually very close between the two training cases, which matches our third insight shown by Theorem \ref{theorem_client_size_relation}. Thus, if the optimal model size in a centralized scenario with the same amount of training data is not known in advance, the optimal model size in a federated scenario can also be estimated based on the average training compute allocated to each client.

\begin{figure}[t]
\centering
% 第一张图
\includegraphics[width=\linewidth]{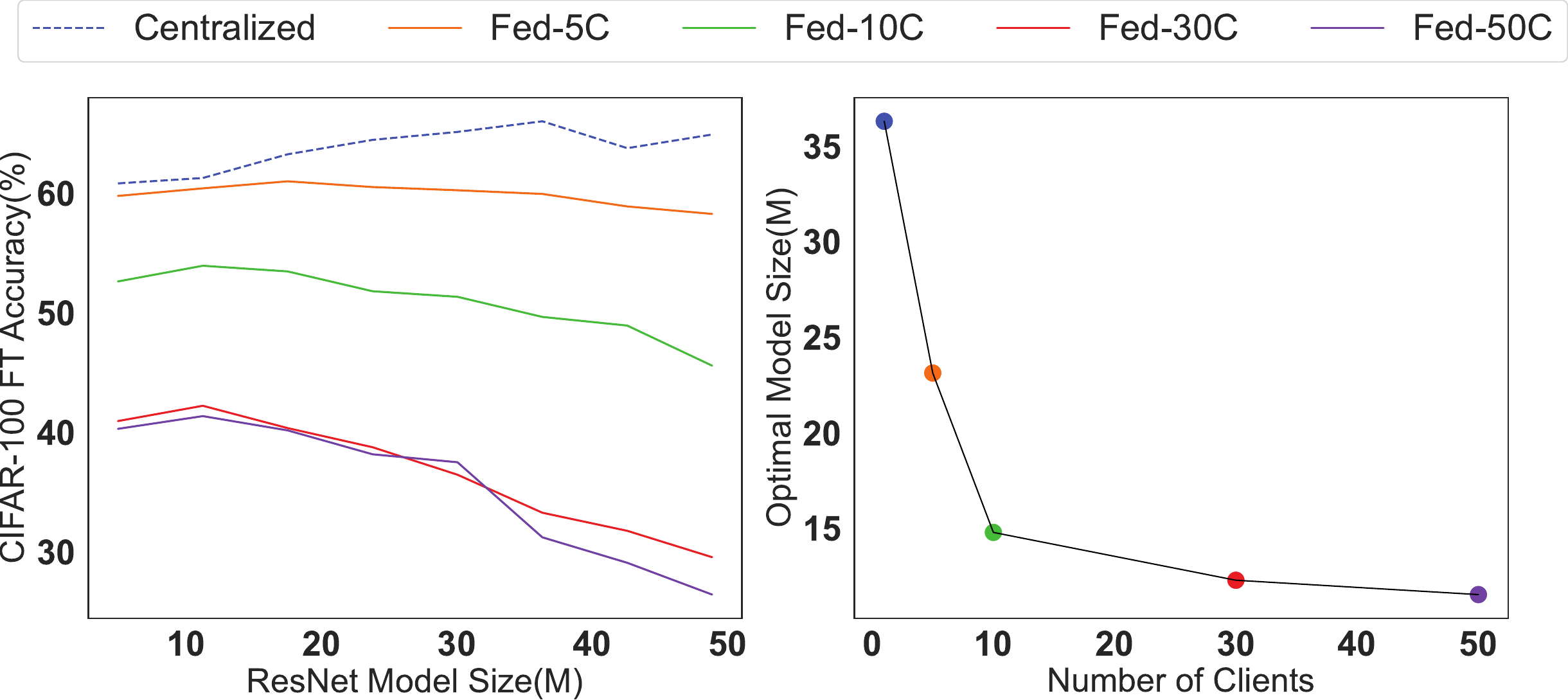}
\caption*{\textbf{(a) Impact of distributed data on the optimal model size.}}

\vspace{0.5em}

% 第二张图
\includegraphics[width=0.8\linewidth]{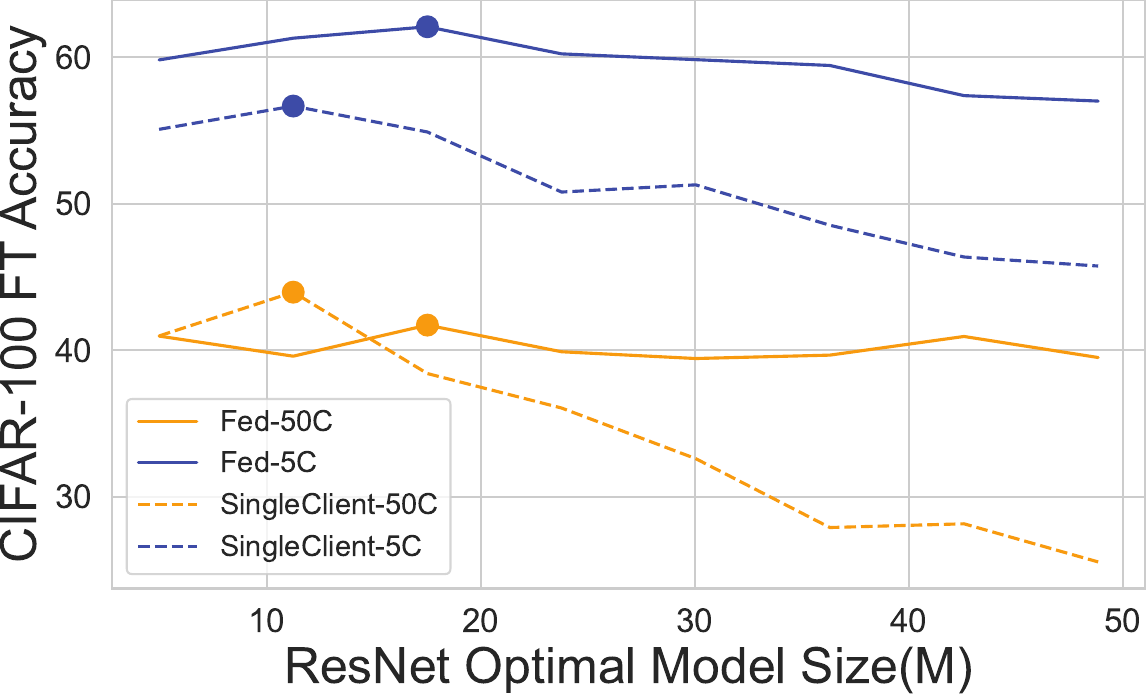}
\caption*{\textbf{(b) Comparison between the optimal model size across all clients and for a single client.}}

\caption{\textbf{Applicability Analysis on ResNets.}}
\label{fig:external_study_resnet}
\end{figure}

\noindent
\textbf{Applicability Study.} Beyond ViTs, we evaluate whether our theoretical insights hold for convolutional models. Figure \ref{fig:external_study_resnet} demonstrates that adapting our theoretical findings to ResNets \cite{he2016deep} exhibits similar behavior on their optimal model sizes, reinforcing that the derived insights are not tied to a specific architecture and may thus serve as general guidelines for size selection in distributed training.

\section{Conclusion}
This work investigates the scaling behavior of large-scale models in federated learning, with a focus on how distributed data affects the estimation of the optimal model size. We derive a PAC-Bayes generalization bound for federated SGD and analyze the global optimal model size under this bound. From this, we obtain three main insights. First, data decentralization reduces the optimal model size, following an approximate negative power law with respect to the number of clients. Second, moving large-scale training to federated settings inevitably lowers the achievable generalization performance. Third, the optimal model size should be estimated by the average training compute per client rather than the total compute or network size. Various experiments on transformer and convolutional models across multiple datasets confirm these findings. We expect our results to offer practical guidance for deploying large-scale models in distributed environments.

\section{Acknowledgments}
This work is partly supported by the Australian Research Council Linkage Project (Grant No.\ LP220200893) and the National Key R\&D Program of China (Grant No.\ 2021YFA1000402).

% The preparation of the \LaTeX{} and Bib\TeX{} files that implement these instructions was supported by Schlumberger Palo Alto Research, AT\&T Bell Laboratories, Morgan Kaufmann Publishers, The Live Oak Press, LLC, and AAAI Press. Bibliography style changes were added by Sunil Issar. \verb+\+pubnote was added by J. Scott Penberthy. George Ferguson added support for printing the AAAI copyright slug. Additional changes to aaai2026.sty and aaai2026.bst have been made by Francisco Cruz, Marc Pujol-Gonzalez, and Mico Loretan.

% \bigskip
% \noindent Thank you for reading these instructions carefully. We look forward to receiving your electronic files!

\bibliography{aaai2026}
\appendix

\onecolumn

\section{Appendix A - Full Proofs for Theoretical Analysis}

\textbf{Problem Setup.} To align with the settings of Federated Learning \cite{mcmahan2017communication}, we assume that the federated scenario consists of $n$ clients and a single central server that connects all clients. Each client $i \in \{1, \ldots, n\}$ stores a local dataset $\mathcal{D}_i$ with an average size $m$ denoted by $m = \frac{1}{n}\sum_{i=1}^n|\mathcal{D}_i|$. Therefore, the size of the total data can be defined as $nm$. Suppose that the training will be repeated for $T$ rounds, the training process at round $j\in \{1, \ldots, T\}$ can be expressed as:
\begin{equation}
\label{eq_aggregate}
\Bar{\theta}_{i}(j) = \frac{1}{n} \sum\limits_{i = 1}^n \theta_{i}(j),
\end{equation}
\begin{equation}
\label{eq_client_train}
\theta_{i}(j+1) = \Bar{\theta}_{i}(j) - \eta \nabla_{\Bar{\theta}_{i}(j)}\mathbb{E}_{\zeta_i \sim \mathcal{D}_i} F(\Bar{\theta}; \zeta_i).
\end{equation} 
where Eq.(\ref{eq_aggregate}) denotes the process of model aggregation and updates to the global model that occurred on the central server, and Eq.(\ref{eq_client_train}) represents the training of the global model using local data $\mathcal{D}_i$ on the client $i$. Furthermore, since training optimization is performed through SGD algorithms, we also define that the batch size for local training is $k_{Fed}m \in \{1, \ldots, m\}$ where $\frac{1}{m} \leq k_{Fed} \leq 1$ and the number of training epochs is $t$. Correspondingly, the baseline centralized scenario holds a dataset $\mathcal{D} = \bigcup_{i=1}^n D_i $ of size $nm$, and the weights of the initial model in this scenario are the same as that in the federated scenario, which is expressed as $\{\theta(0) = \theta_i(0) | i \in n\}$. The training of $\theta$ follows the SGD optimization and will be iterated for $\frac{T}{n}$ rounds to ensure that the total amount of training compute is the same. In each round, the model $\theta$ is trained using data from $\mathcal{D}$ for $t$ epochs with the batch size $k_{Cen}nm \in \{1, \ldots, nm\}$ where $\frac{1}{nm} \leq k_{Cen} \leq 1$. Furthermore, we consistently consider the same batch size and a constant learning rate $\eta$, so we have $k_{Fed}m = k_{Cen}nm$.

\subsection{Preliminaries}

\textbf{Generalization Error.} Formally, considering the hypothesis class of a model is $\Theta \subset \mathbb{R}^d$, machine learning algorithms aim to find the vector of model parameters $\theta \in \Theta $ that minimizes the expected risk $\mathcal{R(\theta)} = \mathbb{E}_{\xi \sim \mathcal{D}} F(\theta; \xi)$ where $d$ is the dimension of the parameter $\theta$, $F$ is the loss function, and $\mathcal{D}$ is the latent distribution of the testing data. Suppose the output parameter $\theta$ follows a distribution $Q$, the expected risk in terms of $Q$ can be formulated as:
\begin{equation}
    \mathcal{R}(Q) = \mathbb{E}_{\theta \sim Q} \mathbb{E}_{\xi \sim \mathcal{D}} F(\theta; \xi).
\end{equation}
In practice, since $\mathcal{D}$ is not known in advance, the expected risk $\mathcal{R}$ is estimated by the empirical risk $\hat{\mathcal{R}}$ in terms of the latent distribution $\hat{\mathcal{D}}$ of the training data, which is defined as:
\begin{equation}
    \hat{\mathcal{R}}(Q) = \mathbb{E}_{\theta \sim Q} \mathbb{E}_{\zeta \sim \hat{\mathcal{D}}} F(\theta; \zeta).
\end{equation}
The difference between $\mathcal{R}$ and $\hat{\mathcal{R}}$ is known as a generalization error.

\vspace{5mm}
\noindent\textbf{PAC-Bayes Upper Bound for Generalization Error.} In the view of PAC-Bayes (Probably Approximately Correct Bayesian) framework \cite{mcallester1998some, mcallester1999pac}, the hypothesis function learned by a stochastic algorithm is sampled randomly from the hypothesis class, and the distance between the distribution of the output hypothesis $Q$ and the prior $P$ (generally assumed to be Gaussian or Uniform distribution) denotes the generalization capability of the algorithm. Therefore, there exists a classic result that uniformly bounding the expected risk $\mathcal{R}(Q)$ as follow:
\begin{lemma}
\label{appendix_lemma_pac_bayes}
For any positive real $\delta \in (0, 1)$, with probability at least $1 - \delta$ over a sample of size $N$, we have the following inequality for all distributions $Q$:
\begin{equation}
\label{appendix_eq_pac_bound}
\begin{aligned}
R(Q) \leq \hat{R}(Q) + \sqrt{\frac{\mathcal{D}(Q||P) + \log(\frac{1}{\delta})+\log(N)+2}{2N - 1}}.
\end{aligned}
\end{equation}
where $\mathcal{D}(Q||P)$ is the KL divergence between the distributions $Q$ and $P$ and is defined as
\begin{equation}
\label{eq_KL_divergence}
\begin{aligned}
\mathcal{D}(Q||P) = \mathbb{E}_{\theta \sim Q} \log(\frac{Q(\theta)}{P(\theta)}).
\end{aligned}
\end{equation}
\end{lemma}

\noindent\textbf{SGD Optimization.} Stochastic Gradient Descent (SGD) is typically used to optimize the empirical risk $\hat{R}$. Assuming that there is a training dataset with size $m$, the mini-batch $\mathcal{S}$ of the training samples is equivalent to a subset of $S$ random indices that are independently and identically (i.i.d.) drawn from the indices set $\{1, \ldots, m\}$. The iteration of SGD can be formally defined as:
\begin{equation}
\label{appendix_eq_SGD_update}
\begin{aligned}
\theta(t+1) &=  \theta(t) - \eta \nabla_{\theta(t)} \hat{\mathcal{R}}(\theta(t)) \\
\quad &= \theta(t) - \eta \frac{1}{S} \sum\limits_{s\in \mathcal{S}} \nabla_{\theta(t)} F_s(\theta(t)).
\end{aligned}
\end{equation}
where $\nabla_{\theta_t} \hat{\mathcal{R}}(\theta_t)$ is the estimated gradient of the empirical risk on the mini-batch and $\eta$ is the learning rate.

\subsection{Proof of Generalization Bound for Federated SGD}

\begin{assumption}
\label{appendix_assum_C}
For Stochastic Gradient Descent (SGD) optimization, since each training sample in the mini-batch is independently and identically sampled from the dataset, all the gradients $\{\nabla_{\theta}F_s(\theta)\}$ computed from individual training sample are uniformly drawn from a Gaussian distribution which centers at the gradient of the expected risk $g(\theta)$ and covariance matrix $C$:
\begin{equation}
\nabla_{\theta}F_s(\theta) \sim \mathcal{N}(g(\theta), C).
\end{equation} 
Thus,  the stochastic gradient $\hat{g}_s(\theta) = \nabla_{\theta(t)}\hat{\mathcal{R}}(\theta(t))$ is also sampled from the Gaussian distribution:
\begin{equation}
\hat{g}_s(\theta) = \frac{1}{S} \sum\limits_{s\in \mathcal{S}} \nabla_{\theta} F_s(\theta) \sim \mathcal{N}(g(\theta), \frac{1}{S} C).
\end{equation} 
Since covariance matrices are (semi) positive-definite, this constant matrix $C$ can be factorized as $C = BB^{\intercal}$. This assumption can be  justified by the central limit theorem if the size of training data is much greater than the batch size. Considering that large models are usually trained on large-scale datasets in practice, this assumption basically holds. 
\end{assumption}

\begin{assumption}
\label{appendix_assum_A}
The loss function $F(\theta)$ in the local region surrounding the minimum is assumed to be convex and 2-order differentiable:
\begin{equation}
F(\theta) = \frac{1}{2} \theta^{\intercal} A \theta.
\end{equation} 
where $A$ is the Hessian matrix surrounding the minimum and is (semi) positive-definite. Without loss of generality, the global minimum of the loss function is also assumed to be 0 when $\theta = 0$. This assumption can be justified when the loss function is smooth and has been shown by empirical experiments in previous studies \cite{li2018visualizing}. 
\end{assumption}

\begin{lemma}
\label{lemma_fedSGD}
Under the above assumptions, if learning rate $\eta$ and batch size $S = k_{Fed}m$ are fixed, we can derive the following analytic solution for the output parameter $\theta_{Fed}(T)$ of federated SGD:
\begin{equation}
\begin{aligned}
&\theta_{Fed}(T) = \frac{1}{n} \sum\limits_{i=1}^{n} \theta_i(T) \\ 
\quad &= \theta_i(0) e^{-T\Bar{A}t} + T \sqrt{\frac{\eta}{k_{Fed}m}} \int_{0}^{t} e^{-T\Bar{A}(t-t')} \Bar{B} dw(t').
\end{aligned}
\end{equation}
where $A_i$ is the Hessian matrix and $B_i$ is the covariance matrix for local training on client $i$, respectively. Besides, we have $\Bar{A} = \frac{1}{n} \sum\limits_{i=1}^{n} A_i$ and $\Bar{B} = \frac{1}{n} \sum\limits_{i=1}^{n} B_i$.
\end{lemma}

\begin{pf}

From the result of the Ornstein-Uhlenbeck process \cite{uhlenbeck1930theory}, the analytical solution for the SGD training with local data from client $i$ in the first round $j=1$ will be

\begin{equation}
\label{eq_training_fed_t1}
\begin{aligned}
        \theta_i(1) &= \theta_i(0) e^{-A_i t} + \sqrt{\frac{\eta}{k_{Fed}m}}\int_{0}^{t} e^{-A_i(t-t')}B_idW(t').
\end{aligned}
\end{equation}

\noindent where $W(t')$ is a white noise and follows $\mathcal{N}(0, I)$. Since local models will be aggregated on the server at each round of federated learning, the analytic solution for local training on client $i$ at the second round $j =2$ should be

\begin{equation}
\label{eq_training_fed_t2}
\begin{aligned}
    \theta_i(2) &= \frac{1}{n} \sum\limits_{i=1}^n \theta_i(1)  e^{-A_i t} + \sqrt{\frac{\eta}{k_{Fed}m}}\int_{0}^{t} e^{-A_i(t-t')}B_idW(t').
\end{aligned}
\end{equation}
Substituting Eq.(\ref{eq_training_fed_t1}) into Eq.(\ref{eq_training_fed_t2}), we have
\begin{equation}
\begin{aligned}
    \theta_i(2) &= \frac{1}{n} \sum\limits_{i=1}^n \left(\theta_i(0) e^{-A_i t} + \sqrt{\frac{\eta}{k_{Fed}m}}\int_{0}^{t} e^{-A_i(t-t')}B_idW(t')\right) e^{-A_i t} \\
    &\quad + \sqrt{\frac{\eta}{k_{Fed}m}}\int_{0}^{t} e^{-A_i(t-t')}B_idW(t') \\
    &= \theta_i(0) e^{-2\Bar{A} t} + \sqrt{\frac{\eta}{k_{Fed}m}}\int_{-t}^{0} e^{-\Bar{A}(t-t')}\bar{B}dW(t') + \sqrt{\frac{\eta}{k_{Fed}m}}\int_{0}^{t} e^{-A_i(t-t')}B_idW(t').
\end{aligned}
\end{equation}
In the same way, we formulate the analytic solution in the round $j=3$ as follows:
\begin{equation}
\begin{aligned}
    \theta_i(3) &= \frac{1}{n} \sum\limits_{i=1}^n (\theta_i(0) e^{-2\Bar{A} t} + \sqrt{\frac{\eta}{k_{Fed}m}}\int_{-t}^{0} e^{-\Bar{A}(t-t')}\bar{B}dW(t')  \\
    &\quad + \sqrt{\frac{\eta}{k_{Fed}m}}\int_{0}^{t} e^{-A_i(t-t')}B_idW(t')) e^{-A_i t} + \sqrt{\frac{\eta}{k_{Fed}m}}\int_{0}^{t} e^{-A_i(t-t')}B_idW(t') \\
    &= \theta_i(0)  e^{-2\Bar{A} t} \frac{1}{n} \sum\limits_{i=1}^n e^{-A_i t} + \sqrt{\frac{\eta}{k_{Fed}m}}\int_{-t}^{0} e^{-\Bar{A}(t-t')}\bar{B}dW(t') \frac{1}{n}\sum\limits_{i=1}^n e^{-A_i t} \\
    &\quad + \sqrt{\frac{\eta}{k_{Fed}m}} \frac{1}{n}\sum\limits_{i=1}^n\int_{0}^{t} e^{-A_i(t-t')} e^{-A_i t} B_idW(t') +  \sqrt{\frac{\eta}{k_{Fed}m}}\int_{0}^{t} e^{-A_i(t-t')}B_idW(t') \\
    &= \theta_i(0)  e^{-3\Bar{A} t} + \sqrt{\frac{\eta}{k_{Fed}m}}\left(\int_{-2t}^{-t} e^{-\Bar{A}(t-t')}\bar{B}dW(t') + \int_{-t}^{0} e^{-\Bar{A}(t-t')}\bar{B}dW(t')\right) \\
    &\quad + \sqrt{\frac{\eta}{k_{Fed}m}}\int_{0}^{t} e^{-A_i(t-t')}B_idW(t')  \\
    &= \theta_i(0)  e^{-3\Bar{A} t} + \sqrt{\frac{\eta}{k_{Fed}m}} \int_{-2t}^{0} e^{-\Bar{A}(t-t')}\bar{B}dW(t') +  \sqrt{\frac{\eta}{k_{Fed}m}}\int_{0}^{t} e^{-A_i(t-t')}B_idW(t').
\end{aligned}
\end{equation}
Similarly, the analytic solution after $T$ rounds of federated training can be derived as the following equation:
\begin{equation}
\label{eq_fed_train}
\begin{aligned}
\theta_{Fed}(T) &=  \frac{1}{n} \sum\limits_{i=1}^{n} \theta_i(T) \\ 
&=  \theta_i(0)  e^{-T\Bar{A} t} + \sqrt{\frac{\eta}{k_{Fed}m}} \int_{(1-T)t}^{0} e^{-\Bar{A}(t-t')}\bar{B}dW(t')  + \sqrt{\frac{\eta}{k_{Fed}m}} \frac{1}{n} \sum\limits_{i=1}^{n} \int_{0}^{t} e^{-A_i(t-t')}B_idW(t') \\
&= \theta_i(0) e^{-T\Bar{A} t} + \sqrt{\frac{\eta}{k_{Fed}m}} \int_{(1-T)t}^{0} e^{-\Bar{A}(t-t')}\bar{B}dW(t') + \sqrt{\frac{\eta}{k_{Fed}m}} \int_{0}^{t} e^{-\Bar{A}(t-t')}\bar{B}dW(t') \\
&= \theta_i(0) e^{-T\Bar{A} t} + \sqrt{\frac{\eta}{k_{Fed}m}} \int_{(1-T)t}^{t} e^{-\Bar{A}(t-t')}\bar{B}dW(t') \\
&= \theta_i(0) e^{-T\Bar{A}t} + \sqrt{\frac{\eta }{k_{Fed}m}} \frac{1 - e^{-T\Bar{A}t}}{\Bar{A}} \Bar{B}  \\ 
&= \theta_0 e^{-T\Bar{A}t} + \sqrt{\frac{\eta }{k_{Fed}m}} \frac{T(1 - e^{-T\Bar{A}t})}{T\Bar{A}} \Bar{B}  \\
&= \theta_0 e^{-T\Bar{A}t} + T \sqrt{\frac{\eta}{k_{Fed}m}} \int_{0}^{t} e^{-T\Bar{A}(t-t')} \Bar{B}  dW(t'). \\
\end{aligned}
\end{equation}
which completes the proof.
\end{pf}
\begin{lemma}
\label{lemma_fed_SGD_property}
Under the Assumption \ref{assum_A}, the stationary distribution of the Ornstein-Uhlenbeck process for the federated SGD,
\begin{equation}
\label{eq_fed_OU_distribution}
\begin{aligned}
q(\theta_{Fed}) = M \exp \left\{-\frac{1}{2} \theta_{Fed}^{\intercal}  \Sigma_{Fed}^{-1} \theta \right\},
\end{aligned}
\end{equation}
has the following property,
\begin{equation}
\label{eq_fed_sigma_property}
\begin{aligned}
T\bar{A} \Sigma_{Fed} + \Sigma_{Fed} T\bar{A} = \frac{T^2\eta}{k_{Fed}m} \Bar{C}.
\end{aligned}
\end{equation}
where $M$ is the normalizer and $\Sigma_{Fed}$ is the covariance matrix of the stationary distribution.
\end{lemma}
\begin{pf} 
From Eq.(\ref{eq_fed_OU_distribution}), we know that
\begin{equation}
\begin{aligned}
\Sigma_{Fed} = \mathbb{E}_{\theta \sim Q}[\theta_{Fed} \theta_{Fed}^{\intercal}].
\end{aligned}
\end{equation}
Then, according to Eq.(\ref{eq_fed_train}), we can derive the following equation:
\begin{equation}
\begin{aligned}
T\Bar{A} \Sigma_{Fed} + \Sigma_{Fed} T\Bar{A} &= \frac{T^2\eta}{k_{Fed}m} \int_{-\infty}^{t} T\Bar{A} e^{-T\Bar{A}(t-t')} \Bar{C} e^{-T\Bar{A}(t-t')} dt' \\
&\quad +  \frac{T^2\eta}{k_{Fed}m} \int_{-\infty}^{t} e^{-T\Bar{A}(t-t')} \Bar{C} e^{-T\Bar{A}(t-t')} dt' T\Bar{A} \\
&= \frac{T^2\eta}{k_{Fed}m}  \int_{-\infty}^{t} \frac{d}{dt'} (e^{-T\Bar{A}(t-t')} \Bar{C} e^{-T\Bar{A}(t-t')})  \\
&= \frac{T^2\eta}{k_{Fed}m} \Bar{C}.
\end{aligned}
\end{equation}
which completes the proof.
\end{pf}
\begin{theorem}
\label{theorem_fed_pac_bound}
For any positive real $\delta \in (0, 1)$, with probability at least $1 - \delta$ over a distributed training data set with a total size $nm$ across all clients, we have the following inequality for the distribution $Q_{Fed}$ of the output hypothesis function of federated SGD:
\begin{equation}
\label{appendix_eq_fed_pac_bound}
\begin{aligned}
& R(Q_{Fed}) \leq  \hat{R}(Q_{Fed}) + \sqrt{\frac{-\log(\det(\Sigma_{Fed})) + \frac{T\eta}{2k_{Fed}m} \text{tr}(\Bar{C}\Bar{A}^{-1})  - d + 2\log(\frac{1}{\delta}) + 2\log (nm) + 4}{4nm - 2}}.
\end{aligned}
\end{equation}
where $d$ is the dimension of the parameter (the model size) and $\text{tr}(\Bar{C}\Bar{A}^{-1})$ is the trace of the product matrix $\Bar{C}\Bar{A}^{-1}$.
\end{theorem}
\begin{pf}
Similarly to the classical Pac-Bayesian framework, we suppose the prior distribution over the parameter space $\theta$ is $P$ and the distribution of the learned hypothesis from federated SGD algorithm is $Q$. Then according to Eq.(\ref{eq_fed_OU_distribution}), the densities of the stationary distribution $Q$ and the prior distribution $P$ are respectively $q(\theta)$ and $p(\theta)$ in terms of the parameter $\theta$ and can be expressed as the following equations:
\begin{equation}
\begin{aligned}
    q(\theta) &= \frac{1}{\sqrt{2\pi \det(\Sigma_{Fed})}} \exp \left \{ - \frac{1}{2}  \theta^{\intercal} \Sigma_{Fed}^{-1} \theta \right \}, \\
    p(\theta) &= \frac{1}{\sqrt{2\pi \det(I)}} \exp \left \{ - \frac{1}{2}  \theta^{\intercal} I \theta \right \}. 
\end{aligned}
\end{equation}
Thus we have
\begin{equation}
\begin{aligned}
\log \left( \frac{q(\theta)}{p(\theta)} \right) &= \log \left( \frac{\sqrt{2\pi \det(I)}}{\sqrt{2\pi \det(\Sigma_{Fed})}} \exp \left\{ \frac{1}{2} \theta^{\intercal} I \theta - \frac{1}{2} \theta^{\intercal} \Sigma_{Fed}^{-1} \theta \right\} \right) \\
&= \frac{1}{2} \log \left( \frac{1}{\det(\Sigma_{Fed})} \right) + \frac{1}{2} \left( \theta^{\intercal} I \theta - \theta^{\intercal} \Sigma_{Fed}^{-1} \theta \right).
\end{aligned}
\end{equation}
Here we can calculate the KL divergence between the distribution $Q$ and $P$ by applying Eq.(\ref{eq_KL_divergence}) in Lemma \ref{appendix_lemma_pac_bayes}:
\begin{equation}
\label{eq_raw_KL}
\begin{aligned}
D(Q||P) &= \mathbb{E}_{\theta \sim Q} \left( \log \frac{Q(\theta)}{P(\theta)} \right) \\
&= \int_{\theta \in \Theta} \log \left( \frac{q(\theta)}{p(\theta)} \right) q(\theta) d\theta \\
&= \int_{\theta \in \Theta} \left[ \frac{1}{2} \log \left( \frac{1}{\det(\Sigma_{Fed})} \right) + \frac{1}{2} \left( \theta^{\intercal} I \theta - \theta^{\intercal} \Sigma_{Fed}^{-1} \theta \right) \right] q(\theta) d\theta \\
&= \frac{1}{2} \log \left( \frac{1}{\sqrt{\det(\Sigma_{Fed})}} \right) + \frac{1}{2} \int_{\theta \in \Theta} \theta^{\intercal} I \theta q(\theta) d\theta - \frac{1}{2} \int_{\mathbb{R}^{|\mathcal{S}|}} \theta^{\intercal} \Sigma_{Fed}^{-1} q(\theta) d\theta \\
&= \frac{1}{2} \log \left( \frac{1}{\sqrt{\det(\Sigma_{Fed})}} \right) + \frac{1}{2} \mathbb{E}_{\theta \sim \mathcal{N}(0,\Sigma_{Fed})} \theta^{\intercal} I \theta - \frac{1}{2} \mathbb{E}_{\theta \sim \mathcal{N}(0,\Sigma_{Fed})} \theta^{\intercal} \Sigma_{Fed}^{-1} \theta \\
&= \frac{1}{2} \log \left( \frac{1}{\sqrt{\det(\Sigma_{Fed})}} \right) + \frac{1}{2} \text{tr}(\Sigma_{Fed} - I).
\end{aligned}
\end{equation}
Since we have proved from Lemma \ref{lemma_fed_SGD_property} that $T\Bar{A} \Sigma_{Fed} + \Sigma_{Fed} T\Bar{A} =  \frac{T^2\eta}{k_{Fed}m} \Bar{C}$, we have
\begin{equation}
\begin{aligned}
\Bar{A} \Sigma_{Fed} \Bar{A}^{-1} + \Sigma_{Fed}    &= \frac{T^2\eta}{Tk_{Fed}m} \Bar{C} \Bar{A}^{-1} \\
\text{tr}(\Bar{A} \Sigma_{Fed} \Bar{A}^{-1} + \Sigma_{Fed} ) &= \text{tr}(\frac{T\eta}{k_{Fed}m} \Bar{C} \Bar{A}^{-1}).
\end{aligned}
\end{equation}
For the left hand side, we can change it to the following equation:
\begin{equation}
\begin{aligned}
\text{LHS} &=  \text{tr}(\Bar{A} \Sigma_{Fed} \Bar{A}^{-1} + \Sigma_{Fed}) \\ 
&=  \text{tr}(\Bar{A} \Sigma_{Fed} \Bar{A}^{-1}) + \text{tr}(\Sigma_{Fed}) \\ 
&=  \text{tr}(\Bar{A} \Bar{A}^{-1} \Sigma_{Fed} ) + \text{tr}(\Sigma_{Fed}) \\ 
&=  \text{tr}(\Sigma_{Fed} ) + \text{tr}(\Sigma_{Fed}) \\ 
&=  2\text{tr}(\Sigma_{Fed} ).
\end{aligned}
\end{equation}
Therefore, 
\begin{equation}
\begin{aligned}
\text{tr}(\Sigma_{Fed} ) &= \frac{1}{2} \text{tr}(\frac{T\eta}{k_{Fed}m} \Bar{C} \Bar{A}^{-1}) = \frac{T\eta}{2k_{Fed}m} \text{tr}(\Bar{C} \Bar{A}^{-1}). \\
\end{aligned}
\end{equation}
On the other side, we can simply calculate that $\text{tr}(I) = d$, because $I \in \mathbb{R}^{d \times d}$, where $d$ is the dimension of the parameter $\theta$. Then we can have 
\begin{equation}
\label{eq_fed_KL_divergence}
\begin{aligned}
D(Q_{Fed}||P) &= - \frac{1}{2} \log(\det(\Sigma_{Fed})) + \frac{1}{2} \text{tr}(\Sigma_{Fed}) - \frac{1}{2} \text{tr}(I) \\
 &= - \frac{1}{2} \log(\det(\Sigma_{Fed})) + \frac{ T\eta}{4k_{Fed}m} \text{tr}( \bar{C} \bar{A}^{-1}) - \frac{1}{2} d. 
\end{aligned}
\end{equation}
By inserting the Eq.(\ref{eq_fed_KL_divergence}) into Eq.(\ref{eq_pac_bound}), we can drive the following inequality for the global training sample set of size $nm$:
\begin{equation}
\begin{aligned}
R(Q_{Fed}) \leq \hat{R}(Q_{Fed}) + \sqrt{\frac{-  \log(\det(\Sigma_{Fed})) + \frac{T\eta}{2k_{Fed}m} \text{tr}( \bar{C} \bar{A}^{-1}) - d  + 2\log(\frac{1}{\delta}) + 2\log (nm) + 4}{4nm - 2}}.
\end{aligned}
\end{equation}
which has completed the proof.
\end{pf}
\begin{assumption}
\label{appendix_assum_symmetric}
Like previous work \cite{he2019control, jastrzkebski2017three}, we study a special case of generalization bound and assume that $A$ and $\Sigma$ are symmetric matrices so that their product satisfies $A\Sigma = \Sigma A$, suggesting the homogeneity between the local geometry around the global minima and the stationary distribution across every dimension in the parameter space.
\end{assumption}
\begin{theorem}
\label{theorem_new_fed_pac_bound}
Under all the assumptions of Theorem \ref{theorem_fed_pac_bound} and with Assumption \ref{appendix_assum_symmetric}, we have the following generalization bound for the stationary distribution of federated SGD:  
\begin{equation}
\label{eq_fed_pac_bound_2}
\begin{aligned}
& R(Q_{Fed}) - \hat{R}(Q_{Fed}) \\ 
&\leq \sqrt{\frac{d\log(\frac{2k_{Fed}m}{T\eta}) - \log(\det(\Bar{C}\Bar{A}^{-1})) + \frac{T\eta}{2k_{Fed}m} \text{tr}(\Bar{C}\Bar{A}^{-1}) - d + 2\log(\frac{1}{\delta}) + 2\log (nm) + 4}{4nm - 2}}.
\end{aligned}
\end{equation}
\end{theorem}

\begin{pf}
Based on Assumption \ref{assum_symmetric}, we can reformulate Eq.(\ref{eq_fed_sigma_property}) in Lemma \ref{lemma_fed_SGD_property} to
\begin{equation}
\label{eq_new_fed_property}
\begin{aligned}
T\Bar{A} \Sigma_{Fed} + \Sigma_{Fed} T\Bar{A} &= \frac{T^2\eta}{k_{Fed}m} \Bar{C} \\ 
2T \Sigma_{Fed} \Bar{A} &= \frac{T^2\eta}{k_{Fed}m}  \Bar{C} \\
\Sigma_{Fed} &= \frac{T\eta}{2k_{Fed}m} \Bar{C}\Bar{A}^{-1}.
\end{aligned}
\end{equation}
By substituting Eq.(\ref{eq_new_fed_property}) into Eq.(\ref{appendix_eq_fed_pac_bound}) and rearranging the equation, we have
\begin{equation}
\begin{aligned}
&R(Q_{Fed}) - \hat{R}(Q_{Fed}) \\ 
&\leq \sqrt{\frac{-\log(\det(\frac{T\eta}{2k_{Fed}m}\Bar{C}\Bar{A}^{-1})) + \frac{T\eta}{2k_{Fed}m} \text{tr}(\bar{C}\bar{A}^{-1}) - d  + 2\log(\frac{1}{\delta}) + 2\log (nm) + 4}{4nm - 2}}\\
&\leq \sqrt{\frac{-\log((\frac{T\eta}{2k_{Fed}m})^d\det(\Bar{C}\Bar{A}^{-1})) + \frac{T\eta}{2k_{Fed}m} \text{tr}(\bar{C}\bar{A}^{-1}) - d  + 2\log(\frac{1}{\delta}) + 2\log (nm) + 4}{4nm - 2}}\\
&\leq \sqrt{\frac{d\log(\frac{2k_{Fed}m}{T\eta}) - \log(\det(\Bar{C}\Bar{A}^{-1})) + \frac{T\eta}{2k_{Fed}m} \text{tr}(\bar{C}\bar{A}^{-1}) - d  + 2\log(\frac{1}{\delta}) + 2\log (nm) + 4}{4nm - 2}}.
\end{aligned}
\end{equation}
which completes the proof.
\end{pf}

\subsection{Proof of First Insight: The Relationship Between Two Optimal Model Sizes}

\begin{lemma}
\label{lemma_cenSGD}
Under all assumptions of Lemma \ref{lemma_fedSGD}, if learning rate $\eta$ and batch size $S = k_{Cen}nm$ are fixed, we can derive the following analytic solution for the output parameter of centralized SGD trained on same amount of training data:
\begin{equation}
\label{eq_cenSGD}
\begin{aligned}
\theta_{Cen}(T) = \theta(0) e^{-\frac{T}{n}At} + \frac{T}{n} \sqrt{\frac{\eta}{k_{Cen}nm}} \int_{0}^{t} e^{-\frac{T}{n}A(t-t')}BdW(t')).
\end{aligned}
\end{equation}
where $A$ is the Hessian matrix and $B$ is the covariance matrix for global training on $nm$ data.
\end{lemma}

\begin{pf}
Based on the result of the Ornstein-Uhlenbeck process \cite{uhlenbeck1930theory}, we can simply derive the following analytic solution for the baseline centralized SGD:

\begin{equation}
\begin{aligned}
\theta_{Cen}(T)  = \theta(0) e^{-\frac{T}{n}At} + \frac{T}{n} \sqrt{\frac{\eta}{k_{Cen}nm}} \int_{0}^{t} e^{-\frac{T}{n}A(t-t')}BdW(t')).
\end{aligned}
\end{equation}
thus completing the proof.
\end{pf}

\begin{lemma}
\label{lemma_cen_SGD_property}
When Assumption \ref{assum_A} holds, the Ornstein-Uhlenbeck process’s stationary distribution for the baseline centralized SGD,
\begin{equation}
\label{eq_cen_OU_distribution}
\begin{aligned}
q(\theta_{Cen}) = M \exp \left\{-\frac{1}{2} \theta^{\intercal}  \Sigma_{Cen}^{-1} \theta \right\},
\end{aligned}
\end{equation}
has the following property,
\begin{equation}
\label{eq_cen_sigma_property}
\begin{aligned}
\frac{T}{n}A \Sigma_{Cen} + \Sigma_{Cen} \frac{T}{n} A = \frac{T^2\eta}{k_{Cen}n^3m} C.
\end{aligned}
\end{equation}

\end{lemma}

\begin{pf}
Based on Eq.(\ref{eq_cen_OU_distribution}), we know that
\begin{equation}
\label{eq_cen_covariance}
\begin{aligned}
\Sigma_{Cen} = \mathbb{E}_{\theta \sim Q}[\theta_{Cen} \theta_{Cen}^{\intercal}].
\end{aligned}
\end{equation}
Then, by combining Eq.(\ref{eq_cenSGD}) and Eq.(\ref{eq_cen_covariance}), we can derive the following equation:
\begin{equation}
\begin{aligned}
\frac{T}{n} A \Sigma_{Cen} + \Sigma_{Cen} \frac{T}{n}A  &= \frac{T^2\eta}{k_{Cen}n^3m} \int_{-\infty}^{t}  \frac{T}{n}A e^{-\frac{T}{n}A(t-t')}Ce^{-\frac{T}{n}A(t-t')}dt' \\
&\quad + \frac{T^2\eta}{k_{Cen}n^3m} \int_{-\infty}^{t} e^{-\frac{T}{n}A(t-t')}Ce^{-\frac{T}{n}A(t-t')}dt' \frac{T}{n}A \\
&= \frac{ T^2\eta}{k_{Cen}n^3m}  \int_{-\infty}^{t} \frac{d}{dt'} (e^{-\frac{T}{n}A(t-t')}Ce^{-\frac{T}{n}A(t-t')}) \\
&= \frac{T^2\eta}{k_{Cen}n^3m} C.
\end{aligned}   
\end{equation}
which completes the proof.
\end{pf}

\begin{lemma}
\label{appendix_lemma_cen_pac_bound}
Under all the assumptions of Theorem \ref{theorem_new_fed_pac_bound}, we have the following generalization bound for the stationary distribution of centralized SGD trained on the same amount of training data:  
\begin{equation}
\label{eq_cen_pac_bound2}
\begin{aligned}
& R(Q_{Cen}) - \hat{R}(Q_{Cen}) \\
&\leq \sqrt{\frac{d\log(\frac{2k_{Cen}n^2m}{T\eta}) - \log(\det(CA^{-1})) + \frac{T\eta}{2k_{Cen}n^2m} \text{tr}(CA^{-1}) - d + 2\log(\frac{1}{\delta}) + 2\log (nm) + 4}{4nm - 2}}.
\end{aligned}
\end{equation}
\end{lemma}

\begin{pf}
Since we have proved from Lemma \ref{lemma_cen_SGD_property} that $\frac{T}{n}A \Sigma_{Cen} + \Sigma_{Cen} \frac{T}{n} A = \frac{T^2\eta}{k_{Cen}n^3m} C$, we have
\begin{equation}
\label{eq_trace_cen_SGD}
\begin{aligned}
A \Sigma_{Cen} + \Sigma_{Cen} A &= \frac{T\eta}{k_{Cen}n^2m} C \\
A \Sigma_{Cen} A^{-1} + \Sigma_{Cen} &= \frac{T\eta}{k_{Cen}n^2m} C A^{-1} \\
\text{tr} (A \Sigma_{Cen} A^{-1} + \Sigma_{Cen}) &= \text{tr}({\frac{T\eta}{k_{Cen}n^2m} C A^{-1}}) \\
2\text{tr} (\Sigma_{Cen}) &= \text{tr}({\frac{T\eta}{k_{Cen}n^2m} C A^{-1}}) \\
\text{tr} (\Sigma_{Cen}) &= \frac{T\eta}{2k_{Cen}n^2m} \text{tr}({ C A^{-1}}).
\end{aligned}
\end{equation}
Similarly to the proof of Theorem \ref{theorem_fed_pac_bound}, by substituting the Eq.(\ref{eq_trace_cen_SGD}) into Eq.(\ref{eq_raw_KL}), we can compute the KL divergence between the distribution of the output hypothesis and the prior as below:
\begin{equation}
\label{eq_cen_KL_divergence}
\begin{aligned}
D(Q_{Cen}||P) &= - \frac{1}{2} \log(\det(\Sigma_{Cen})) + \frac{1}{2} \text{tr}(\Sigma_{Cen}) - \frac{1}{2} \text{tr}(I) \\
 &= - \frac{1}{2} \log(\det(\Sigma_{Cen})) + \frac{ T\eta}{4k_{Cen}n^2m} \text{tr}( \bar{C} \bar{A}^{-1}) - \frac{1}{2} d. 
\end{aligned}
\end{equation}
According to Lemma \ref{appendix_lemma_pac_bayes}, then we can derive the following inequality to bound the generalization bound error of the baseline centralized SGD:
\begin{equation}
\label{eq_cen_pac_bound}
\begin{aligned}
R(Q_{Cen}) \leq \hat{R}(Q_{Cen}) + \sqrt{\frac{-  \log(\det(\Sigma_{Cen})) + \frac{T\eta}{2k_{Cen}n^2 m} \text{tr}(CA^{-1}) - d  + 2\log(\frac{1}{\delta}) + 2\log (nm) + 4}{4nm - 2}}.
\end{aligned}
\end{equation}
Since we have assumed that $A \Sigma = \Sigma A$ from Assumption \ref{assum_symmetric}, we can reformulate Eq.(\ref{eq_cen_sigma_property}) to
\begin{equation}
\label{eq_cen_new_property}
\begin{aligned}
\frac{T}{n}A \Sigma_{Cen} + \Sigma_{Cen} \frac{T}{n} A  &= \frac{T^2\eta}{k_{Cen}n^3m} C \\
2 \Sigma_{Cen} A &= \frac{T\eta}{k_{Cen}n^2m}  C \\
\Sigma_{Cen} &= \frac{T\eta}{2k_{Cen}n^2m} CA^{-1}.
\end{aligned}
\end{equation}
By inserting  Eq.(\ref{eq_cen_new_property}) into Eq.(\ref{eq_cen_pac_bound}) and rearranging the equation, we have
\begin{equation}
\begin{aligned}
&R(Q_{Cen}) - \hat{R}(Q_{Cen}) \\
&\leq \sqrt{\frac{-\log(\det(\frac{T\eta}{2k_{Cen}n^2m}CA^{-1})) + \frac{T\eta}{2k_{Cen}n^2m} \text{tr}(CA^{-1}) - d  + 2\log(\frac{1}{\delta}) + 2\log (nm) + 4}{4nm - 2}} \\
&\leq \sqrt{\frac{d\log(\frac{2k_{Cen}n^2m}{T\eta}) - \log(\det(CA^{-1})) + \frac{T\eta}{2k_{Cen}n^2m} \text{tr}(CA^{-1}) - d  + 2\log(\frac{1}{\delta}) + 2\log (nm) + 4}{4nm - 2}}.
\end{aligned}
\end{equation}
The proof has been completed.
\end{pf}

\begin{assumption}
\label{appendix_assump_fair_comp}
Under a fair comparison that the distribution and size of training data are the same for both training, we assume that the average data distribution $\Bar{\mathcal{D}}$ across all clients with average size $m$ is independently and identically (i.i.d.) drawn from the central dataset $\mathcal{D}$ with size $nm$ and the following properties are satisfied:
\begin{equation}
\Bar{A} \approx A, 
\end{equation}
\begin{equation}
\text{tr}(\Bar{C}) \approx \frac{1}{n^{\gamma}} \text{tr} (C).
\end{equation}
where $\gamma$ is a constant that $\gamma > 1$. The trace of the covariance matrix is equivalent to the sum of loss gradient variance across each dimension of the parameter. This assumption can be justified by the central limit theorem when the average size $m$ of the local dataset is large enough.
\end{assumption}

\begin{theorem}
\label{appendix_theorem_size_relation}
When all the above assumptions hold, the optimal model size under the output hypothesis function of
federated SGD has the following analytic solution:
\begin{equation}
\label{appendix_eq_optimal_size_fed}
d_{Fed}^* = \frac{-4n\log((\det(\Bar{C}\Bar{A}^{-1})) + (\frac{4nT\eta}{k_{Fed}m} - \frac{T\eta}{k_{Fed}m^2}) \text{tr}(\Bar{C}\Bar{A}^{-1})) + 8n\log(\frac{1}{\delta}) + 8n\log(nm) - \frac{4}{m} + 8n}{8n - \frac{2}{m} - 4n\log(\frac{2k_{Fed}m}{T\eta})}.
\end{equation}
and the optimal model size for centralized SGD has the following analytic solution:
\begin{equation}
\label{appendix_eq_optimal_size_cen}
d_{Cen}^* = \frac{-4n\log((\det(CA^{-1})) + (\frac{4T\eta}{k_{Cen}nm} - \frac{T\eta}{k_{Cen}n^2m^2}) \text{tr}(CA^{-1})) + 8n\log(\frac{1}{\delta}) + 8n\log(nm) - \frac{4}{m} + 8n}{8n - \frac{2}{m} - 4n\log(\frac{2k_{Cen}n^2m}{T\eta})}.
\end{equation}
Considering the same batch size $k_{Fed}m = k_{Cen}nm$, these solutions suggest that:
\begin{equation}
\lim_{T \to \infty} d_{Fed}^* = \frac{\rho}{n^{\gamma - 1}}d_{Cen}^*, 
\end{equation}
where
\begin{equation}
\rho = \frac{4m-1}{4m -\frac{1}{n}}>1.
\end{equation}
\end{theorem}

\begin{pf}
At the beginning, we define
\begin{equation}
\begin{aligned}
L_{Fed} &= \frac{d\log(\frac{2k_{Fed}m}{T\eta}) - \log(\det(\Bar{C}\Bar{A}^{-1})) + \frac{T\eta}{2k_{Fed}m} \text{tr}(\Bar{C}\Bar{A}^{-1})) - d  + 2\log(\frac{1}{\delta}) + 2\log (nm) + 4}{4nm-2},  \\
L_{Cen} &=\frac{d\log(\frac{2k_{Cen}n^2m}{T\eta}) - \log(\det(CA^{-1})) + \frac{T\eta}{2k_{Cen}n^2m} \text{tr}(CA^{-1}) - d  + 2\log(\frac{1}{\delta}) + 2\log (nm) + 4}{4nm-2}. 
\end{aligned}
\end{equation}
Then, Eq.(\ref{eq_fed_pac_bound_2}) and Eq.(\ref{eq_cen_pac_bound2}) can be turned into

\begin{equation}
\begin{aligned}
R(Q_{Fed}) &\leq \hat{R}(Q_{Fed}) + \sqrt{L_{Fed}}, \\
R(Q_{Cen}) &\leq \hat{R}(Q_{Cen}) + \sqrt{L_{Cen}}.
\end{aligned}
\end{equation}
To find the optimal model size that minimizes the generalization bound, we start by calculating the derivative of $L_{Fed}$ with respect to the average amount of training data $m$ on clients as follow:
\begin{equation}
\begin{aligned}
\frac{\partial L_{Fed}}{\partial m} = \frac{G_1 - G_2}{(4nm-2)^2}.
\end{aligned}
\end{equation}
where
\begin{equation}
\begin{aligned}
G_1 &= (4nm-2)(\frac{d+2}{m} - \frac{T\eta}{2k_{Fed}m^2}\text{tr}(\bar{C}\bar{A}^{-1})), \\
G_2 &= (4n)(d\log(\frac{2k_{Fed}m}{T\eta})  - \log(\det(\bar{C}\bar{A}^{-1})) + \frac{T\eta}{2k_{Fed}m}\text{tr}(\bar{C}\bar{A}^{-1}) -d + 2\log(\frac{1}{\sigma}) + 2\log(nm) +4).
\end{aligned}
\end{equation}
By setting $\frac{\partial L_{Fed}}{\partial m} = 0$, we derive the following optimal model size:
\begin{equation}
\begin{aligned}
& (4nm-2)\frac{d}{m} + (4nm-2)(\frac{2}{m} - \frac{T\eta}{2k_{Fed}m^2}\text{tr}(\bar{C}\bar{A}^{-1})) \\
&= 4n(d\log(\frac{2k_{Fed}m}{T\eta}) - \log(\det(\bar{C}\bar{A}^{-1}))  + \frac{T\eta}{2k_{Fed}m}\text{tr}(\bar{C}\bar{A}^{-1}))  -d + 2\log(\frac{1}{\sigma}) + 2\log(nm) +4) \\
&(4n - \frac{2}{m} - 4n\log(\frac{2k_{Fed}m}{T\eta}) + 4n)d \\
&= 4n(-\log(\det(\bar{C}\bar{A}^{-1})) + \frac{T\eta}{2k_{Fed}m}\text{tr}(\bar{C}\bar{A}^{-1}) + 2\log(\frac{1}{\sigma}) + 2\log(nm) +4) \\
&\quad - (4nm-2)(\frac{2}{m} - \frac{T\eta}{2k_{Fed}m^2}\text{tr}(\bar{C}\bar{A}^{-1})) \\
& d^{*}_{Fed} = \frac{-4n\log((\det(\Bar{C}\Bar{A}^{-1})) + (\frac{4nT\eta}{k_{Fed}m} - \frac{T\eta}{k_{Fed}m^2}) \text{tr}(\Bar{C}\Bar{A}^{-1})) + 8n\log(\frac{1}{\delta}) + 8n\log(nm) - \frac{4}{m} + 8n}{8n - \frac{2}{m} - 4n\log(\frac{2k_{Fed}m}{T\eta})}.
\end{aligned}
\end{equation}
In the same way, we obtain the below optimal model size $d_{Cen}^*$ for the baseline centralized SGD:

\begin{equation}
\begin{aligned}
d_{Cen}^* =\frac{-4n\log \left( \det \left( CA^{-1} \right) \right) +\left( \frac{4T\eta}{k_{Cen}nm}-\frac{T\eta}{k_{Cen}n^2m^2} \right) \text{tr}\left( CA^{-1} \right) +8n\log \left( \frac{1}{\delta} \right) +8n\log \left( nm \right) -\frac{4}{m}+8n}{\left( 8n-\frac{2}{m}-4n\log \left( \frac{2k_{Cen}n^2m}{T\eta} \right) \right)}.
\end{aligned}
\end{equation}
When $T\rightarrow \infty$ and based on Assumption \ref{appendix_assump_fair_comp}, we have
\begin{equation}
\begin{aligned}
\underset{T\rightarrow \infty}{\lim}\frac{d_{Fed}^*}{d_{Cen}^*}&=\frac{\left( \frac{4nT\eta}{k_{Fed}m}-\frac{T\eta}{k_{Fed}m^2} \right) \text{tr}\left( \bar{C}\bar{A}^{-1} \right)}{-4n\log \left( \frac{2k_{Fed}m}{T\eta} \right)}\times \frac{-4n\log \left( \frac{2k_{Cen}n^2m}{T\eta} \right)}{\left( \frac{4T\eta}{k_{Cen}nm}-\frac{T\eta}{k_{Cen}n^2m^2} \right) \text{tr}\left( CA^{-1} \right)}\\
&=\frac{\left( \frac{4n\eta}{k_{Fed}m}-\frac{\eta}{k_{Fed}m^2} \right) \text{tr}\left( \bar{C}\bar{A}^{-1} \right) \log \left( \frac{2k_{Cen}n^2m}{T\eta} \right)}{\left( \frac{4\eta}{k_{Cen}nm}-\frac{\eta}{k_{Cen}n^2m^2} \right) \text{tr}\left( CA^{-1} \right) \log \left( \frac{2k_{Fed}m}{T\eta} \right)}\\
&=\frac{\left( \frac{4\eta}{k_{Cen}m}-\frac{\eta}{k_{Cen}m^2} \right) \frac{1}{n^{\gamma}}}{\frac{4\eta}{k_{Cen}nm}-\frac{\eta}{k_{Cen}n^2m^2}}\\
&=\frac{4n^2m-n^2}{4n^{\gamma +1}m-n^{\gamma}}.
\end{aligned}
\end{equation}
This can be simplified to give
\begin{equation}
\begin{aligned}
\underset{T\rightarrow \infty}{\lim}d_{Fed}^*&=\frac{4n^2m-n^2}{4n^{\gamma +1}m-n^{\gamma}}d_{Cen}^*\\
&=\frac{4m-1}{4n^{\gamma -1}m-n^{\gamma -2}}d_{Cen}^*\\
&=\frac{4m-1}{4m-n^{-1}}\frac{1}{n^{\gamma -1}}d_{Cen}^*\\
&=\frac{\rho}{n^{\gamma -1}}d_{Cen}^*.
\end{aligned}
\end{equation}
where $ \rho = \frac{4m-1}{4m -\frac{1}{n}}>1$. The proof has been completed.
\end{pf}

\subsection{Proof of Second Insight: Theoretical Evidence for Generalization Gap}

\begin{lemma}
\label{lemmma_error_diff}
When all the above assumptions hold, based on the results of the generalization bound and the optimal model size under this bound, the difference in the optimal generalization error between federated SGD and centralized SGD is given by the following equation:
\begin{equation}
\label{eq_final_generalization_error}
\mathcal{G}_{Fed}^* - \mathcal{G}_{Cen}^* = \frac{\text{tr}(CA^{-1})\left(\frac{T\eta}{2k_{Fed}n^{\gamma}m} \log(\frac{2k_{Fed}n^{\gamma}m}{T\eta}) - \frac{T\eta}{2k_{Cen}n^2m} \log(\frac{2k_{Cen}n^2m}{T\eta})\right)}{4nm-2}.
\end{equation}
where $\mathcal{G}^*$ is the optimal generalization error computed with the optimal model size $d^*$.
\end{lemma}

\begin{pf}
When $T \to \infty$, we can observe that the optimal model sizes  shown by Eq.(\ref{appendix_eq_optimal_size_fed}) and Eq.(\ref{appendix_eq_optimal_size_cen}) will turn to the following equations:
\begin{equation}
\label{eq_optimal_model_size_Fed_limit}
\begin{aligned}
\underset{T\rightarrow \infty}{\lim}d^*_{Fed}&=\frac{\left( \frac{4nT\eta}{k_{Fed}m}-\frac{T\eta}{k_{Fed}m^2} \right) \text{tr}\left( \bar{C}\bar{A}^{-1} \right)}{-4n\log \left( \frac{2k_{Fed}m}{T\eta} \right)} \\
&=\frac{\left( \frac{4nT\eta}{k_{Fed}m}-\frac{T\eta}{k_{Fed}m^2} \right) \text{tr}\left( \bar{C}\bar{A}^{-1} \right)}{4n}\\
&=\frac{\left( \frac{\left( 4nm-1 \right) T\eta}{k_{Fed}m^2} \right) \text{tr}\left( \bar{C}\bar{A}^{-1} \right)}{4n}\\
&=\frac{\left( 4nm-1 \right)}{2nm}\left( \frac{T\eta}{2k_{Fed}m} \right) \text{tr}\left( \bar{C}\bar{A}^{-1} \right), 
\end{aligned}
\end{equation}
\begin{equation}
\label{eq_optimal_model_size_Cen_limit}
\begin{aligned}
\underset{T\rightarrow \infty}{\lim}d^*_{Cen}&=\frac{\left( \frac{4T\eta}{k_{Cen}nm}-\frac{T\eta}{k_{Cen}n^2m^2} \right) \text{tr}\left( CA^{-1} \right)}{-4n\log \left( \frac{2k_{Cen}n^2m}{T\eta} \right)}\\
&=\frac{\left( \frac{4T\eta}{k_{Cen}nm}-\frac{T\eta}{k_{Cen}n^2m^2} \right) \text{tr}\left( CA^{-1} \right)}{4n}\\
&=\frac{\left( 4nm-1 \right)}{2nm}\left( \frac{T\eta}{2k_{Cen}n^2m} \right) \text{tr}\left( CA^{-1} \right). 
\end{aligned}
\end{equation}
There is a common term $\frac{4nm-1}{2nm}$ in both Eq.(\ref{eq_optimal_model_size_Fed_limit}) and Eq.(\ref{eq_optimal_model_size_Cen_limit}).
For the ease of the following computation, we simplify both the Eq.(\ref{eq_optimal_model_size_Fed_limit}) and Eq.(\ref{eq_optimal_model_size_Cen_limit}) into equations:
\begin{equation}
\label{eq_optimal_model_size_Fed_simplify}
\begin{aligned}
\underset{T\rightarrow \infty}{\lim}d^*_{Fed}
&\approx \left( \frac{T\eta}{2k_{Fed}m} \right) \text{tr}\left( \bar{C}\bar{A}^{-1} \right), 
\end{aligned}
\end{equation}
\begin{equation}
\label{eq_optimal_model_size_Cen_simplify}
\begin{aligned}
\underset{T\rightarrow \infty}{\lim}d_{Cen}^*
&\approx\left( \frac{T\eta}{2k_{Cen}n^2m} \right) \text{tr}\left( CA^{-1} \right).
\end{aligned}
\end{equation}
Therefore, the optimal generalization bound for federated learning based on the optimal model size can be formulated as:
\begin{equation}
\begin{aligned}
\lim_{T \to \infty} \left[R\left( Q_{Fed} \right) -\hat{R}\left( Q_{Fed} \right) \right] &\leq \sqrt{L_{Fed}}\\
&\leq \mathcal{G}_{Fed}^*\\
\end{aligned}
\end{equation}
where 
\begin{equation}
\label{eq_g_error_fed_start}
\begin{aligned}
&\mathcal{G}_{Fed}^* = \frac{-\log(\det(\Sigma_{Fed})) +\frac{T\eta}{2k_{Fed}m}\text{tr}\left( \bar{C}\bar{A}^{-1} \right) -d^*_{Fed}+2\log \left( \frac{1}{\delta} \right) +2\log \left( nm \right) +4}{4nm-2} \\
&= \frac{d^*_{Fed}\log \left( \frac{2k_{Fed}m}{T\eta} \right) -\log \left( \det \left( \bar{C}\bar{A}^{-1} \right) \right) +\frac{T\eta}{2k_{Fed}m}\text{tr}\left( \bar{C}\bar{A}^{-1} \right) -d^*_{_{Fed}}+2\log \left( \frac{1}{\delta} \right) +2\log \left( nm \right) +4}{4nm-2}.
\end{aligned}
\end{equation}
Similarly, the optimal generalization bound for centralized learning based on the optimal model size is defined as:
\begin{equation}
\label{eq_g_error_cen_start}
\begin{aligned}
& \mathcal{G}_{Cen}^* = \frac{-\log(\det(\Sigma_{Cen})) +\frac{T\eta}{2k_{Cen}n^2m}\text{tr}\left( CA^{-1} \right) -d^*_{Cen}+2\log \left( \frac{1}{\delta} \right) +2\log \left( nm \right) +4}{4nm-2} \\
&= \frac{d^*_{Cen}\log \left( \frac{2k_{Cen}n^2m}{T\eta} \right) -\log \left( \det \left( CA^{-1} \right) \right) +\frac{T\eta}{2k_{Cen}n^2m}\text{tr}\left( CA^{-1} \right) -d^*_{Cen}+2\log \left( \frac{1}{\delta} \right) +2\log \left( nm \right) +4}{4nm-2}.
\end{aligned}
\end{equation}
Then, we have
\begin{equation}
\label{eq_g_error_diff}
\begin{aligned}
\mathcal{G}_{Fed}^* - \mathcal{G}_{Cen}^* &= \frac{-\log(\det(\Sigma_{Fed})) +\frac{T\eta}{2k_{Fed}m}\text{tr}\left( \bar{C}\bar{A}^{-1} \right) -d^*_{_{Fed}}}{4nm-2} \\
&\quad - \frac{-\log(\det(\Sigma_{Cen})) +\frac{T\eta}{2k_{Cen}n^2m}\text{tr}\left( CA^{-1} \right) -d^*_{Cen}}{4nm-2} \\
&= \frac{d^*_{Fed}\log \left( \frac{2k_{Fed}m}{T\eta} \right) -\log \left( \det \left( \bar{C}\bar{A}^{-1} \right) \right) +\frac{T\eta}{2k_{Fed}m}\text{tr}\left( \bar{C}\bar{A}^{-1} \right) -d^*_{_{Fed}}}{4nm-2} \\
&\quad - \frac{d^*_{Cen}\log \left( \frac{2k_{Cen}n^2m}{T\eta} \right) -\log \left( \det \left( CA^{-1} \right) \right)  +\frac{T\eta}{2k_{Cen}n^2m}\text{tr}\left( CA^{-1} \right) -d^*_{Cen}}{4nm-2}
\end{aligned}
\end{equation}
Since Assumption \ref{appendix_assump_fair_comp} implies that $\text{tr}(\bar{C}\bar{A}^{-1}) = \frac{1}{n^\gamma}\text{tr}(CA^{-1})$ and $\Sigma_{Fed} = \frac{1}{n^\gamma}\Sigma_{Cen}$, by inserting Eq.(\ref{eq_optimal_model_size_Fed_limit}) and Eq.(\ref{eq_optimal_model_size_Cen_limit}),  the Eq.(\ref{eq_g_error_diff}) can be simplified into
\begin{equation}
\begin{aligned}
\mathcal{G}_{Fed}^* - \mathcal{G}_{Cen}^* &= \frac{-\log(\det(\Sigma_{Fed})) +\frac{T\eta}{2k_{Fed}m}\text{tr}\left( \bar{C}\bar{A}^{-1} \right) - \frac{T\eta}{2k_{Fed}m} \text{tr}\left( \bar{C}\bar{A}^{-1}\right)}{4nm-2} \\
&\quad - \frac{-\log(\det(\Sigma_{Cen})) +\frac{T\eta}{2k_{Cen}n^2m}\text{tr}\left( CA^{-1} \right) - \frac{T\eta}{2k_{Cen}n^2m}  \text{tr}\left( CA^{-1} \right)}{4nm-2} \\
&= \frac{\frac{T\eta}{2k_{Fed}n^\gamma m} \text{tr}( CA^{-1}) \log \left( \frac{2k_{Fed}n^\gamma m}{T\eta} \right) -\log \left( \det \left( CA^{-1} \right) \right)}{4nm-2} \\
&\quad - \frac{\frac{T\eta}{2k_{Cen}n^2m} \text{tr}( CA^{-1}) \log \left( \frac{2k_{Cen}n^2 m}{T\eta} \right) -\log \left( \det \left( CA^{-1} \right) \right)}{4nm-2} \\
&= \frac{\text{tr}(CA^{-1})\left(\frac{T\eta}{2k_{Fed}n^{\gamma}m} \log(\frac{2k_{Fed}n^{\gamma}m}{T\eta}) - \frac{T\eta}{2k_{Cen}n^2m} \log(\frac{2k_{Cen}n^2m}{T\eta})\right)}{4nm-2}.
\end{aligned}
\end{equation}
which completes the proof.
\end{pf}

\begin{theorem}
\label{appendix_theorem_gap}
When all the above assumptions hold, from the results of the generalization bound and the optimal model size under this bound, we have the following inequality for the optimal generalization error of federated SGD and centralized SGD using the same amount of training compute:
\begin{equation}
\mathcal{G}_{Fed}^* - \mathcal{G}_{Cen}^* \geq 0. 
\end{equation}
when $n$ satisfies the below property:
\begin{equation}
\quad n \geq \sqrt[\gamma - 1]{\frac{e\eta}{2k_{Cen}nm}}.
\end{equation}
\end{theorem}

\begin{pf}
If we define $f(x)=\frac{1}{x}\log(x)$, then Eq.(\ref{eq_final_generalization_error}) can be turned into 
\begin{equation}
\begin{aligned}
\mathcal{G}_{Fed}^* - \mathcal{G}_{Cen}^* &= \frac{\text{tr}(CA^{-1})\left(\frac{1}{x_1}\log(x_1) - \frac{1}{x_2} \log(x_2)\right)}{4nm-2}.
\end{aligned}
\end{equation}
where
\begin{equation}
\begin{aligned}
    x_1 &= \frac{2k_{Fed}n^{\gamma}m}{T\eta} \\
    x_2 &= \frac{2k_{Cen}n^2m}{T\eta} \\
\end{aligned}
\end{equation}
Due to the monotonicity of $f(x)=\frac{1}{x}\log(x)$ and the equal batch size $k_{Fed}m = k_{Cen}nm$, we have
\begin{equation}
\left( \frac{T\eta}{2n^{\gamma +1}k_{Cen}m}\log \left( \frac{2n^{\gamma +1}k_{Cen}m}{T\eta} \right)-\frac{T\eta}{2k_{Cen}n^2m}\log \left( \frac{2k_{Cen}n^2m}{T\eta} \right) \right) 
\ge 0,
\end{equation}
when
\begin{equation}
\label{eq_condition_e}
\frac{2n^{\gamma+1}k_{Cen}m}{T\eta}\leq e.
\end{equation}
By rearranging Eq.(\ref{eq_condition_e}), we can also have
\begin{equation}
    T\ge \frac{2n^{\gamma +1}k_{Cen}m}{e\eta}.
\end{equation}
On the other hand, we know that the centralized training will be iterated for $\frac{T}{n}$ steps from the problem setup, which implies that:
\begin{equation}
\begin{aligned}
    \frac{T}{n} &\geq 1 \\
    T &\geq n.
\end{aligned}
\end{equation}
In summary, there are
\begin{equation}
\left\{ \begin{array}{l}
	T\ge \frac{2n^{\gamma +1}k_{Cen}m}{e\eta}\\
	T\ge n
\end{array} \right. .
\end{equation}
By combing the two conditions, we have
\begin{equation}
\frac{2n^{\gamma +1}k_{Cen}m}{e\eta}\ge n.
\end{equation}
Solving for this gives
\begin{equation}
\label{eq_n_condition}
n\ge \sqrt[\gamma -1]{\frac{e\eta}{2k_{Cen}nm}}.
\end{equation}
Since we have assumed that both $C$ and $A$ are (semi) positive-definite in Assumption \ref{assum_C} and \ref{assum_A}, we can simply derive
\begin{equation}
\text{tr}(CA^{-1}) \geq 0.
\end{equation}
Therefore, when the condition in Eq.(\ref{eq_n_condition}) holds, we have
\begin{equation}
\mathcal{G}_{Fed}^* - \mathcal{G}_{Cen}^* = \frac{\text{tr}(CA^{-1})\left(\frac{T\eta}{2k_{Fed}n^{\gamma}m} \log(\frac{2k_{Fed}n^{\gamma}m}{T\eta}) - \frac{T\eta}{2k_{Cen}n^2m} \log(\frac{2k_{Cen}n^2m}{T\eta})\right)}{4nm-2} \geq 0.
\end{equation}
where $nm$ is the total size of the data and is assumed to be a very large value that satisfies $4nm-2>0$. The proof has been completed.
\end{pf}

\subsection{Proof of Third Insight: Estimating Optimal Model Size by Average Training Compute Between Clients}

\begin{assumption}
\label{assump_bound_C}
(Bounded Inter-client Gradient Variance) There exists $\psi \geq 0$ such that:
\begin{equation}
    \frac{1}{n}\sum\limits_{i=1}^{n}||C_i - \Bar{C}||^2 \leq \psi^2.
\end{equation}
\end{assumption}
\begin{assumption}
\label{assump_bound_A}
(Bounded Inter-client Hessian Variance) The loss function for any client $i$ follows:
\begin{equation}
    \frac{1}{n}\sum\limits_{i=1}^{n}||A_i - \Bar{A}||^2 \leq \tau.
\end{equation}
\end{assumption}

\begin{lemma}
\label{lemma_single_client_SGD}
Under all the assumptions of Lemma \ref{lemma_fedSGD}, if learning rate $\eta$ and batch size $S = k_im$ are fixed, we can derive the following analytic solution for the local output parameter $\Acute{\theta}_{i}(T)$ on client $i$:
\begin{equation}
\begin{aligned}
\Acute{\theta}_{i}(T) &= \theta_i(0) e^{-TA_it} + T \sqrt{\frac{\eta}{k_{i}m}} \int_{0}^{t} e^{-TA_i(t-t')}B_idW(t')).
\end{aligned}
\end{equation} 
\end{lemma}

\begin{pf}
Based on Eq.(\ref{eq_cenSGD}), we can simply derive the following analytic solution for this baseline training that only uses the local data on client $i$ and is also iterated for $T$ rounds:
\begin{equation}
\begin{aligned}
\Acute{\theta}_{i}(T) &= \theta_i(0) e^{-TA_it} + T \sqrt{\frac{\eta}{k_{i}m}} \int_{0}^{t} e^{-TA_i(t-t')}B_idW(t')).
\end{aligned}
\end{equation} 
which completes the proof.
\end{pf}

\begin{lemma}
\label{lemma_single_client_bound}
Under all the assumptions of Theorem \ref{theorem_new_fed_pac_bound}, we have the following generalization bound for the stationary distribution of  SGD training with sorely the local data on client $i$ under the same number of training compute:  
\begin{equation}
\label{appendix_eq_single_client_pac_bound}
\begin{aligned}
R(Q_{i}) - \hat{R}(Q_{i}) \leq \sqrt{\frac{d_i\log(\frac{2k_{i}m}{T\eta}) - \log(\det(C_iA_i^{-1})) + \frac{T\eta}{2k_{i}m} \text{tr}(C_iA_i^{-1}) - d_i + 2\log(\frac{1}{\delta}) + 2\log (m) + 4}{4m - 2}}.
\end{aligned}
\end{equation}
\end{lemma}
\begin{pf}
Similarly to the proof of Lemma \ref{appendix_lemma_cen_pac_bound}, we first derive
\begin{equation}
\label{eq_single_client_property}
\begin{aligned}
\Sigma_{i} &= \frac{T\eta}{2k_{i}m} C_iA_i^{-1}.
\end{aligned}
\end{equation}
Then, we reformulate the Eq.(\ref{eq_cen_pac_bound2}) in terms of Eq.(\ref{eq_single_client_property}) and the size $m$ of local data on client $i$ to obtain the following generalization bound:
\begin{equation}
\label{eq_single_client_pac_bound}
\begin{aligned}
R(Q_{i}) - \hat{R}(Q_{i}) \leq \sqrt{\frac{d_i\log(\frac{2k_{i}m}{T\eta}) - \log(\det(C_iA_i^{-1})) + \frac{T\eta}{2k_{i}m} \text{tr}(C_iA_i^{-1}) - d_i + 2\log(\frac{1}{\delta}) + 2\log (m) + 4}{4m - 2}}.
\end{aligned}
\end{equation}
which completes the proof.
\end{pf}

\begin{theorem}
\label{appendix_theorem_client_size_relation}
When Assumption \ref{assump_bound_C} and \ref{assump_bound_A} both hold, under all the assumptions of Theorem \ref{theorem_new_fed_pac_bound}, and considering the same batch size $\{k_{Fed}m = k_{i}m | i \in n\}$, the optimal model size at the client level has the following analytic solution:
\begin{equation}
d_{i}^* = \frac{ -4\log((\det(C_iA_i^{-1})) + (\frac{4T\eta}{k_im} - \frac{T\eta}{k_im^2}) \text{tr}(C_iA_i^{-1})) + 8\log(\frac{1}{\delta}) + 8\log(m) - \frac{4}{m} + 8}{8 - \frac{2}{m} - 4\log(\frac{2k_{i}m}{T\eta})}.
\end{equation}
and also suggests that:
\begin{equation}
\lim_{T\to\infty}d_{Fed}^* \approx \frac{1}{n} \sum\limits_{i=1}^n d_i^* .
\end{equation}
\end{theorem}

\begin{pf}
We firstly define 
\begin{equation}
\begin{aligned}
L_{i} &= 
\frac{d\log \left( \frac{2k_im}{T\eta} \right) -\log \left( \det \left( C_iA_{i}^{-1} \right) \right) +\frac{T\eta}{2k_im}\text{tr}\left( C_iA_{i}^{-1} \right) -d+2\log \left( \frac{1}{\delta} \right) +2\log \left( m \right) +4}{4m-2}.
\end{aligned}
\end{equation}
Then, Eq.(\ref{eq_single_client_pac_bound}) becomes
\begin{equation}
\begin{aligned}
R(Q_{i}) &\leq \hat{R}(Q_{i}) + \sqrt{L_{i}}.
\end{aligned}
\end{equation}
We calculate the derivative of $L_{i}$ with respect to the amount of training data $m$ on the client $i$ as follow:
\begin{equation}
\begin{aligned}
\frac{\partial L_{i}}{\partial m} &= 
\frac{G_1-G_2}{\left( 4m-2 \right) ^2}.
\end{aligned}
\end{equation}
where
\begin{equation}
\begin{aligned}
&G_1=\left( 4m-2 \right) \left( \frac{d+2}{m}-\frac{T\eta}{2k_im^2}\text{tr}\left( C_iA_{i}^{-1} \right) \right), \\
&G_2=4\left( d\log \left( \frac{2k_im}{T\eta} \right) -\log \left( \det \left( C_iA_{i}^{-1} \right) \right) +\frac{T\eta}{2k_im}\text{tr}\left( C_iA_{i}^{-1} \right) -d+2\log \left( \frac{1}{\delta} \right) +2\log \left( m \right) +4 \right). 
\end{aligned}
\end{equation}
Similarly, to find the optimal size, we set $\frac{\partial L_{i}}{\partial m} = 0$ and derive
\begin{equation}
\label{eq_optimal_model_size_1c}
\begin{aligned}
&\left( 4m-2 \right) \left( \frac{d+2}{m}-\frac{T\eta}{2k_im^2}\text{tr}\left( C_iA_{i}^{-1} \right) \right) \\
&=4\left( d\log \left( \frac{2k_im}{T\eta} \right) -\log \left( \det \left( C_iA_{i}^{-1} \right) \right) +\frac{T\eta}{2k_im}\text{tr}\left( C_iA_{i}^{-1} \right) -d+2\log \left( \frac{1}{\delta} \right) +2\log \left( m \right) +4 \right)\\
&d_i^*=
\frac{-4\log \left( \det \left( C_iA_{i}^{-1} \right) \right) +\left( \frac{4T\eta}{k_im}-\frac{T\eta}{k_im^2} \right) \text{tr}\left( C_iA_{i}^{-1} \right) +8\log \left( \frac{1}{\delta} \right) +8\log \left( m \right) -\frac{4}{m}+8}{8-\frac{2}{m}-4\log \left( \frac{2k_{Fed}m}{T\eta} \right)}.
\end{aligned}
\end{equation}
In optimal model size Eq.(\ref{appendix_eq_optimal_size_fed}), all except for the term $(\frac{4nT\eta}{k_{Fed}m} - \frac{T\eta}{k_{Fed}m^2})\text{tr}(\Bar{C}\bar{A}^{-1})$ are bounded by the $\log$ operator. While in Eq.(\ref{eq_optimal_model_size_1c}), there is a similar term $(\frac{4T\eta}{k_im}-\frac{T\eta}{k_im^2})\text{tr}(C_iA_i^{-1})$. Since we considered using the same batch size, we have
\begin{equation}
\begin{aligned}
k_{Fed}m  &= k_{i}m.
\end{aligned}
\end{equation}
By applying Assumption \ref{assump_bound_C} and Assumption \ref{assump_bound_A}, we derive the below result when $T \to \infty$:
\begin{equation}
\begin{aligned}
\underset{T\rightarrow \infty}{\lim}\frac{d_{Fed}^*}{\frac{1}{n}\sum_{i=1}^n{d_i^*}}&=\frac{\left( \frac{4nT\eta}{k_{Fed}m}-\frac{T\eta}{k_{Fed}m^2} \right) \text{tr}\left( \bar{C}\bar{A}^{-1} \right)}{-4n\log \left( \frac{2k_{Fed}m}{T\eta} \right)}\times {\frac{-4\log \left( \frac{2k_{Fed}m}{T\eta} \right)}{\left( \frac{4T\eta}{k_{Fed}m}-\frac{T\eta}{k_{Fed}m^2} \right) \frac{1}{n}\sum_{i=1}^n \text{tr}\left( C_iA_{i}^{-1} \right)}}\\
&\approx \frac{1\left( \frac{4n}{k_{Fed}m}-\frac{1}{k_{Fed}m^2} \right)}{n\left( \frac{4}{k_{Fed}m}-\frac{1}{k_{Fed}m^2} \right)}\\
&=\frac{4m-\frac{1}{n}}{4m-1}.
\end{aligned}
\end{equation}
Because $
\ 1\le \frac{4m-\frac{1}{n}}{4m-1}\le \frac{4m}{4m-1}\approx 1$, we have
\begin{equation}
\underset{T\rightarrow \infty}{\lim}d_{Fed}^*\approx \frac{1}{n}\sum_{i=1}^n{d_i^*}.
\end{equation}
which completes the proof.
\end{pf}

\newpage

\section{Appendix B - Experiment Setup}
In this section, we have provided two tables to present our experiment setup. Table \ref{tab1} shows the details about the experiment system, which include the specific settings for the model architecture, dataset, scenario and training. Table \ref{tab2} demonstrates the setup of the running environment, including the configuration of our test server.

\begin{table}[h!]
\centering
\caption{System Settings.}
\label{tab1}
\vskip 0.15in
\begin{tabular}{l c} \hline
System & Value \\ [0.5ex] 
\hline
Model Architecture & Vision Transformer \cite{dosovitskiy2020image} \\
Pre-train Method & Masked Auto-encoder \cite{he2022masked} \\
Pre-train Dataset & Mini-ImageNet \cite{vinyals2016matching} \\
Fine-tune Dataset & CIFAR-100 \cite{krizhevsky2009learning}, ImageNet \cite{deng2009imagenet} \\
Total Pre-Training Compute & 900, 000 \\
Number of Clients in Federated Scenario & $\{3, 5, 7, 10, 20, 30, 40, 50\}$ \\
Data Distribution on Clients & I.I.D \\
Model Size Options (Millions) & $\{11.62, 18.71, 25.80, 32.89, 39.97,$\\
& $47.06, 54.15, 61.24, 68.33, 75.41\}$ \\
Fine-tuning Epochs & $\{1, 5\}$ \\ 
Batch Size & 1024 \\
Base Learning Rate & 1.5e-4\\
\hline
\end{tabular}
\vskip -0.1in
\end{table}

\begin{table}[h!]
\centering
\caption{Running Environment Settings.}
\label{tab2}
\begin{tabular}{l c} \hline
Config & Details \\ [0.5ex] 
 \hline
 Server GPU Count & 8 \\
 Server GPU Type & RTX A5000 (24GB) \\ 
 Server CPU Type & AMD EPYC 7513 32-core \\
 CUDA & 11.3 \\
 Framework & PyTorch \\ \hline
\end{tabular}
\end{table}

\end{document}